\documentclass[sigconf]{acmart}
\usepackage{amsmath,amssymb,amsfonts}

\usepackage{algorithmic}
\usepackage{graphicx}
\usepackage{xspace}
\usepackage{textcomp}
\usepackage{xcolor}
\usepackage{adjustbox}  
\usepackage{algorithm}
\usepackage{algorithmic}
\usepackage{multirow}
\usepackage{booktabs}
\usepackage{color}
\usepackage{subcaption}

\usepackage[para]{footmisc} 

\AtBeginDocument{%
  }

\copyrightyear{2024} 
\acmYear{2024} 
\setcopyright{acmlicensed}\acmConference[DAC '24]{61st ACM/IEEE Design Automation Conference}{June 23--27, 2024}{San Francisco, CA, USA}
\acmBooktitle{61st ACM/IEEE Design Automation Conference (DAC '24), June 23--27, 2024, San Francisco, CA, USA}
\acmDOI{10.1145/3649329.3657380}
\acmISBN{979-8-4007-0601-1/24/06}

\begin{document}
\newcommand{\comt}[1]{{\color{pink} reviewer comments: #1}}
\newcommand{\WJwrite}[1]{{\color{purple} WJ: #1}}
\newcommand{\YYwrite}[1]{{\color{red} YY: #1}}
\newcommand{\ZHwrite}[1]{{\color{brown} ZH: #1}}
\newcommand{\Wei}[1]{{\color{blue} Todo: #1}}
\newcommand{\yytodo}[1]{{\color{magenta} @Yaoyuan plz do: #1}}
\newcommand{\stutodo}[1]{{\color{green} @Coders: plz do: #1}}

\newcommand{\ours}{KATO\xspace}
\newcommand{\tl}{Transfer Learning\xspace}

\def\Eqref#1{Eq.~\eqref{#1}}
\def\Algref#1{Algorithm~\ref{#1}}
\def\Figref#1{Fig.~\ref{#1}}

\newcommand{\var}{{\rm Var}}
\newcommand{\Tr}{^{\rm Tr}}
\newcommand{\vtrans}[2]{{#1}^{(#2)}}
\newcommand{\kron}{\otimes}
\newcommand{\schur}[2]{({#1} | {#2})}
\newcommand{\schurdet}[2]{\left| ({#1} | {#2}) \right|}
\newcommand{\had}{\circ}
\newcommand{\diag}{{\rm diag}}
\newcommand{\invdiag}{\diag^{-1}}
\newcommand{\rank}{{\rm rank}}
\newcommand{\nullsp}{{\rm null}}
\newcommand{\tr}{{\rm tr}}
\renewcommand{\vec}{{\rm vec}}
\newcommand{\vech}{{\rm vech}}
\renewcommand{\det}[1]{\left| #1 \right|}
\newcommand{\pdet}[1]{\left| #1 \right|_{+}}
\newcommand{\pinv}[1]{#1^{+}}
\newcommand{\erf}{{\rm erf}}
\newcommand{\hypergeom}[2]{{}_{#1}F_{#2}}

\renewcommand{\a}{{\bf a}}
\renewcommand{\b}{{\bf b}}
\renewcommand{\c}{{\bf c}}
\renewcommand{\d}{{\rm d}}  
\newcommand{\e}{{\bf e}}
\newcommand{\f}{{\bf f}}
\newcommand{\g}{{\bf g}}
\newcommand{\h}{{\bf h}}
\renewcommand{\k}{{\bf k}}
\newcommand{\m}{{\bf m}}
\newcommand{\n}{{\bf n}}
\renewcommand{\o}{{\bf o}}
\newcommand{\p}{{\bf p}}
\newcommand{\q}{{\bf q}}
\renewcommand{\r}{{\bf r}}
\newcommand{\s}{{\bf s}}
\renewcommand{\t}{{\bf t}}
\renewcommand{\u}{{\bf u}}
\renewcommand{\v}{{\bf v}}
\newcommand{\w}{{\bf w}}
\newcommand{\x}{{\bf x}}
\newcommand{\y}{{\bf y}}
\newcommand{\z}{{\bf z}}
\newcommand{\A}{{\bf A}}
\newcommand{\B}{{\bf B}}
\newcommand{\D}{{\bf D}}
\newcommand{\E}{{\bf E}}
\newcommand{\F}{{\bf F}}
\renewcommand{\H}{{\bf H}}
\newcommand{\I}{{\bf I}}
\newcommand{\J}{{\bf J}}
\newcommand{\K}{{\bf K}}
\renewcommand{\L}{{\bf L}}
\newcommand{\M}{{\bf M}}
\newcommand{\N}{\mathcal{N}}  
\newcommand{\MN}{\mathcal{MN}} 
\newcommand{\Acal}{\mathcal{A}}
\newcommand{\Ocal}{\mathcal{O}}
\newcommand{\Dcal}{\mathcal{D}}
\newcommand{\Ycal}{\mathcal{Y}}
\newcommand{\Zcal}{\mathcal{Z}}
\newcommand{\Fcal}{\mathcal{F}}
\newcommand{\Vcal}{\mathcal{V}}
\newcommand{\Lcal}{\mathcal{L}}
\newcommand{\Tcal}{\mathcal{T}}
\newcommand{\Gcal}{\mathcal{G}}
\newcommand{\Hcal}{\mathcal{H}}
\newcommand{\Scal}{\mathcal{S}}
\newcommand{\Xcal}{\mathcal{X}}

\renewcommand{\O}{{\bf O}}
\renewcommand{\P}{{\bf P}}
\newcommand{\Q}{{\bf Q}}
\newcommand{\R}{{\bf R}}
\renewcommand{\S}{{\bf S}}
\newcommand{\T}{{\bf T}}
\newcommand{\V}{{\bf V}}
\newcommand{\W}{{\bf W}}
\newcommand{\X}{{\bf X}}
\newcommand{\Y}{{\bf Y}}
\newcommand{\Z}{{\bf Z}}
\newcommand{\Mcal}{{\mathcal{M}}}
\newcommand{\Wcal}{{\mathcal{W}}}
\newcommand{\Ucal}{{\mathcal{U}}}

\newcommand{\bfLambda}{\boldsymbol{\Lambda}}

\newcommand{\bsigma}{\boldsymbol{\sigma}}
\newcommand{\balpha}{\boldsymbol{\alpha}}
\newcommand{\bpsi}{\boldsymbol{\psi}}
\newcommand{\bphi}{\boldsymbol{\phi}}
\newcommand{\boldeta}{\boldsymbol{\eta}}
\newcommand{\Beta}{\boldsymbol{\eta}}
\newcommand{\btau}{\boldsymbol{\tau}}
\newcommand{\bvarphi}{\boldsymbol{\varphi}}
\newcommand{\bzeta}{\boldsymbol{\zeta}}
\newcommand{\bepsilon}{\boldsymbol{\epsilon}}

\newcommand{\blambda}{\boldsymbol{\lambda}}
\newcommand{\bLambda}{\mathbf{\Lambda}}
\newcommand{\bOmega}{\mathbf{\Omega}}
\newcommand{\bomega}{\mathbf{\omega}}
\newcommand{\bPi}{\mathbf{\Pi}}

\newcommand{\btheta}{\boldsymbol{\theta}}
\newcommand{\bpi}{\boldsymbol{\pi}}
\newcommand{\bxi}{\boldsymbol{\xi}}
\newcommand{\bSigma}{\boldsymbol{\Sigma}}

\newcommand{\bgamma}{\boldsymbol{\gamma}}
\newcommand{\bGamma}{\mathbf{\Gamma}}

\newcommand{\bmu}{\boldsymbol{\mu}}
\newcommand{\1}{{\bf 1}}
\newcommand{\0}{{\bf 0}}

\newcommand{\bs}{\backslash}

 \newcommand{\notS}{{\backslash S}}
 \newcommand{\nots}{{\backslash s}}
 \newcommand{\noti}{{\backslash i}}
 \newcommand{\notj}{{\backslash j}}
 \newcommand{\nott}{\backslash t}
 \newcommand{\notone}{{\backslash 1}}
 \newcommand{\nottp}{\backslash t+1}

\newcommand{\notk}{{^{\backslash k}}}
\newcommand{\notij}{{^{\backslash i,j}}}
\newcommand{\notg}{{^{\backslash g}}}
\newcommand{\wnoti}{{_{\w}^{\backslash i}}}
\newcommand{\wnotg}{{_{\w}^{\backslash g}}}
\newcommand{\vnotij}{{_{\v}^{\backslash i,j}}}
\newcommand{\vnotg}{{_{\v}^{\backslash g}}}
\newcommand{\half}{\frac{1}{2}}
\newcommand{\msgb}{m_{t \leftarrow t+1}}
\newcommand{\msgf}{m_{t \rightarrow t+1}}
\newcommand{\msgfp}{m_{t-1 \rightarrow t}}

\newcommand{\proj}[1]{{\rm proj}\negmedspace\left[#1\right]}
\newcommand{\argmin}{\operatornamewithlimits{argmin}}
\newcommand{\argmax}{\operatornamewithlimits{argmax}}

\newcommand{\dif}{\dfrac{\dfrac{\mathrm}{den}}{den}{d}}
\newcommand{\abs}[1]{\lvert#1\rvert}
\newcommand{\norm}[1]{\lVert#1\rVert}

\newcommand{\ie}{{i.e.,}\xspace}
\newcommand{\eg}{{e.g.,}\xspace}
\newcommand{\etc}{{etc.}\xspace}

\newcommand{\EE}{\mathbb{E}}
\newcommand{\dr}[1]{\nabla #1}
\newcommand{\VV}{\mathbb{V}}
\newcommand{\sbr}[1]{\left[#1\right]}
\newcommand{\rbr}[1]{\left(#1\right)}
\newcommand{\cmt}[1]{}

\newcommand{\bi}{{\bf i}}
\newcommand{\bj}{{\bf j}}
\newcommand{\bK}{{\bf K}}
\newcommand{\Vtr}{\mathrm{Vec}}

\newcommand{\cov}{{\rm Cov}}	

\newtheorem{Proposition}{Proposition}
\newtheorem{Lemma}{Lemma}
\newtheorem{Corollary}{Corollary}
\newtheorem{Remark}{Remark}
\newtheorem{Assumption}{Assumption}
\newtheorem{Property}{Property}

\newcommand{\RR}{\mathbb{R}}
\newcommand{\KL}{\mathrm{KL}}

\newcommand{\bPsi}{\boldsymbol{\Psi}}
\newcommand{\bXi}{\boldsymbol{\Xi}}
\newcommand{\btx}{\textbf{\textit{x}}}
\newcommand{\bty}{\textbf{\textit{y}}}
\newcommand{\btz}{\textbf{\textit{z}}}
\newcommand{\btk}{\textbf{\textit{k}}}

\newcommand{\bupsilon}{\boldsymbol{\upsilon}}

\newcommand{\GP}{\mathcal{GP}}
\newcommand{\TGP}{\mathcal{TGP}}
\newcommand{\TNcal}{\mathcal{TN}}

\title{KATO: Knowledge Alignment And Transfer for Transistor Sizing Of Different Design and Technology}

\settopmatter{printacmref=false}
\author{Wei W. Xing$^{2}$, Weijian Fan$^{1,3}$, Zhuohua	Liu$^{3}$,Yuan Yao$^{4}$and Yuanqi Hu$^{4*}$}
\affiliation{
$^{1}$ Eastern Institute of Technology, Ningbo, China
$^{2}$ The University of Sheffield, U.K.
\\
$^{3}$ College of Mechatronics and Control Engineering, Shenzhen University, Shenzhen, China\\
$^{4}$ School of Integrated Circuit Science and Engineering, Beihang University, Beijing, China \\}


\thanks{$^{*}$ Corresponding author.}
\thanks{This work is supported by NSFC U23A20352
and 62271020.
}

\begin{abstract}
Automatic transistor sizing in circuit design continues to be a formidable challenge.  Despite that Bayesian optimization (BO) has achieved significant success, it is circuit-specific, limiting the accumulation and transfer of design knowledge for broader applications.  This paper proposes (1) efficient automatic kernel construction, (2) the first transfer learning across different circuits and technology nodes for BO, and (3) a selective transfer learning scheme to ensure only useful knowledge is utilized.  These three novel components are integrated into BO with Multi-objective Acquisition Ensemble (MACE) to form Knowledge Alignment and Transfer Optimization (\ours) to deliver state-of-the-art performance: up to 2x simulation reduction and 1.2x design improvement over the baselines.
\vspace{-0.15in}
\end{abstract}

\vspace{-0.3in}

\makeatletter 
\gdef\@ACM@checkaffil{} 
\makeatother

\maketitle

\vspace{-0.1in}

\section{Introduction}
\vspace{-0.05in}
The ongoing miniaturization of circuit devices within integrated circuits (ICs) introduces increasing complexity to the design process, primarily due to the amplified impact of parasitics that cannot be overlooked. In contrast to digital circuit design, where the use of standardized cells and electronic design automation (EDA) tools streamlines the process, analog and mixed-signal (AMS) designs still predominantly rely on the nuanced expertise of designers~\cite{elfadel2019machine}. 
This reliance on analog design experts is not only costly but also introduces the risk of inconsistency and bias. Such challenges can lead to inadequate exploration of the design space and extended design times. As a consequence, it often results in sub-optimal design in terms of power, performance, and area (PPA), underscoring the need for more efficient and objective design methodologies.

Analog circuit design involves two main steps: designing the circuit topology and sizing the transistors, where the former is normally guided by expert knowledge and the latter is labor-intensive and time-consuming. Thus, the transistor sizing is normally solved by algorithms to allow designers to dedicate their expertise more effectively toward the nuanced aspects of topology selection.

Machine learning advancements have heightened interest in the application of reinforcement learning (RL) to transistor sizing.
Specifically, RL agents are trained to optimize the transistor sizing process by rewarding the agent with better designs that deliver higher performance.
\citet{wang2020gcn} introduce a GCN-RL circuit designer that utilizes RL to transfer knowledge efficiently across various technology nodes and topologies with a graph convolutional network (GCN) capturing circuit topology, resulting in enhanced Figures of Merit (FOM) for a range of circuits. \citet{li2021circuit} expand on this by incorporating a circuit attention network and a stochastic method to diminish layout effects, while \citet{settaluri2021automated} utilize a multi-agent approach for sub-block tuning for a larger circuit sizing task.

While RL's promise is clear, it has the following limitations:
(1) the inherent design for complex Markov decision processes of RL makes it an overly complex solution for the relatively straightforward optimization task of transistor sizing, leading to great demand for data of designs and simulations;
(2) RL's complexity and high computational cost may be excessive for smaller teams with limited resources;
and (3) the trained model might contain commercially sensitive design information that can be exploited by competitors.

Bayesian optimization (BO) stands in contrast to reinforcement learning (RL), being a tried-and-tested optimization method for intricate EDA tasks. Its popularity stems from its efficiency, stability, and the fact that it doesn't require pretraining, making it more accessible for immediate deployment.
To solve the transistor sizing challenge as a straightforward optimization problem, \citet{lyu2018EfficientBayesiana} introduce weighted expected improvement (EI), transforming the constrained optimization of transistor sizing into a single-objective optimization task. Building upon this, \citet{lyu2018batch} presents the Multiple Acquisition Function Ensemble (MACE) to facilitate massive parallel simulation, which is essential to harness the power of cluster computing. To resolve the large-scale transistor sizing challenge, \citet{touloupas2021local} implement multiple local Gaussian Processes (GP) with GPU acceleration to significantly improve BO scalability.
The choice of kernel function is pivotal in BO's performance. As a remedy, \citet{bai2021boom} propose deep kernel learning (DKL) to construct automatic GP for efficient design space exploration.

Recognizing the similarities in transistor sizing across different technology nodes, \citet{zhang2022fast} utilize Gaussian Copula to correlate technology nodes with multi-objective BO, aiming to improve the search of the Pareto frontier.

\cmt{
In contrast to RL, Bayesian optimization (BO) is a well-established optimization method for complex engineering and EDA tasks, prized for its efficiency, stability, and ease of implementation without pretraining.
\cite{lyu2018EfficientBayesiana} propose the weight acquisition function, WEIBO, which is a weighted combination of the expected improvement (EI) of the objective and probability of feasibility (PoF) to convert the constrained optimization of transistor sizing into a single objective optimization problem.
\cite{lyu2018batch} further extend this idea with multiple acquisition function ensemble (MACE) to enable massive parallelization that is critical for large-scale transistor sizing.
\cite{touloupas2021local} instead resolve large-scale transistor sizing by introducing multiple local GPs with GPU acceleration. 
Choosing a proper kernel function is critical for the performance of BO. To resolve this challenge, \cite{bai2021boom} shows the use of deep learning for deep kernel learning (DKL) to construct a flexible GP in BO for EDA design space exploitation.
To harness the fact that transistor sizing has similarity under different technology nodes, \cite{wang2022fast} propose Gaussian Copula to establish the correlation of the implemented technology and multi-objective BO to efficiently achieve a higher quality of Pareto frontier.
}

Despite its notable achievements, Bayesian Optimization (BO) faces several challenges that impede its full potential in the application of transistor sizing:
(1) DKL is powerful but data-intensive and demands meticulous neural network design and training;
(2) \tl in BO is only possible for technology and impossible for different designs,
(3) \tl in practice may degrade performance, it is unclear when to use \tl and when not to use \tl.
We propose Knowledge Alignment and Transfer Optimization, \ours, to resolve these challenges simultaneously. The novelty of this work includes:

\vspace{-0.1em}
\begin{enumerate}
  \item As far as the authors are aware, \ours is the first BO sizing method that can transfer knowledge across different circuit designs and technology nodes simultaneously.
  \item To ensure a positive \tl, we propose a simple yet effective Bayesian selection strategy in the BO pipeline.
  \item As an alternative to DKL, we propose a Neural kernel (Neuk), which is more powerful and stable for BO.
  \item \ours is validated using practical analog designs with state-of-the-art methods on multiple experiment setups, showcasing a 2x speedup and 1.2x performance enhancement. 
\end{enumerate}
  
\vspace{-0.15in}
\section{Background}
\vspace{-0.05in}
\subsection{Problem Definition}
\vspace{-0.05in}
\label{sec:problem_definition}
Transistor sizing is typically formulated as a constrained optimization problem. The goal is to maximize a specific performance while ensuring that various metrics meet predefined constraints, \ie 
\begin{equation}
  \label{eq:constrained_optimization}
  \argmax f_0(\x) \quad \text{s.t.} \quad f_i(\x) \geq C_{i},  \quad \forall i \in {1, \ldots, {{N}_{c}}}.
\end{equation}

Here, \(\x \in \RR^d\) represents the design variable, \(f_i(\x)\) calculates the i-th performance metric, and \(C_{i}\) denotes the required minimum value for that metric.
Given the complexity of solving this constrained optimization problem, many convert it into an unconstrained optimization problem by defining a Figure of Merit (FOM) that combines the performance metrics. For instance, \cite{wang2020gcn} defines 
\begin{equation}
  \label{eq: FOM}
  FOM(\x) = \sum_{i=0}^{N_c} w_i\times\frac{min(f_{i}(\x),f_{i}^{bound})-f_{i}^{min}}{f_{i}^{max}-f_{i}^{min}},
\end{equation}
where $f_{i}^{bound}$ is the pre-determined limit and $f_{i}^{max}$ and $f_{i}^{min}$ are the maximum and minimum values obtained through 10,000 random samples, $w_i$ is $-1 or 1$ depending on whether the metric is to be maximized or minimized.
The FOM method is less desirable due to the difficulty in setting all hyperparameters properly.

\cmt
{Denote $\x=[x^{(1)},x^{(2)},\cdots,x^{(D)}]^T \in \Xcal$ as the variation process parameter, and $\Xcal$ the variation parameter space. 
$\Xcal$ is generally a high-dimensional space (\ie large $D$);
Each variable in $\x$ denotes the variation parameters of a circuit during manufacturing, \eg length or width of PMOS and NMOS transistors. 
In general, $\x$ are considered mutually independent Gaussian distributed, 
\begin{equation}
     p(\x) = (2\pi)^{\frac{D}{2}}\exp \left(- ||\x||^2 /2 \right).
\end{equation}

Given a specific value of $\x$, we can evaluate the circuit performance $\y$ (\eg  memory read/write time and amplifier gain) through SPICE simulation, $ \y=\f(\x)$,

where $\f(\cdot)$ is the SPICE simulator, which is considered an expensive and time-consuming black-box function;
$\y=[y^{(1)},y^{(2)},\cdots,y^{(K)}]^T$ are the collections of circuit performance based on the simulations.
When $K$ metrics are all smaller than or equal to their respective thresholds (predefined by designers) $\boldsymbol{t}$, \ie $y^{(k)} \leq t^{(k)}$ for $k=1,\cdots,K$, the circuit is considered as a successful design; otherwise, it is a failure one.

We use failure indicator $I(\x)$, which is 1 if $\x$ leads to a failure design and 0 otherwise, to denote the failure status of a circuit.

Finally, the ground-truth failure rate $\hat{P}_f $ is:

\begin{equation}
    \hat{P}_f = \int_\mathcal{X} I\left( \x \right) p(\x) d\x.
\end{equation}}

\cmt{
Let $\x=[x^{(1)},x^{(2)},\cdots,x^{(D)}]^T \in \Xcal$ as the parameters denote the variation variables, with $\Xcal$ representing the parameter space for such variations. Typically, $\Xcal$ is a high-dimensional space, denoted as $D$, where each element within the vector $\x$ signifies a specific manufacturing-related parameter affecting a circuit, such as the dimensions (length or width) of PMOS and NMOS transistors.
In the context of our analysis, we make a general assumption that the elements of $\x$ are statistically independent and follow a Gaussian distribution:
\begin{equation}
p(\x) = (2\pi)^{\frac{D}{2}}\exp \left(- \frac{||\x||^2}{2} \right).
\end{equation}

Given a particular configuration of $\x$, we can determine the performance of the circuit, denoted as $\y$ (e.g., metrics like memory read/write time and amplifier gain), using SPICE simulation, which is known to be a computationally intensive and time-consuming black-box function. We express this relationship as $\y = \f(\x)$, where $\f(\cdot)$ represents the SPICE simulator.

The vector $\y=[y^{(1)},y^{(2)},\cdots,y^{(K)}]^T$ encompasses the circuit performance metrics obtained from simulations. The success or failure of a circuit design is determined by comparing each metric with its corresponding predefined threshold, $\boldsymbol{t}$, such that $y^{(k)} \leq t^{(k)}$ for $k=1,\cdots,K$. A design is deemed successful if all K metrics meet their respective thresholds; otherwise, it is considered a failure.
To represent the failure status of a circuit, we employ the binary indicator function $I(\x)$, which equals 1 when $\x$ leads to a failure in the design and 0 otherwise:
\begin{equation}
    I(\x) = 
    \begin{cases}
 1 & \text{ if } y^{(k)} > t^{(k)}  \\
 0 & \text{ if } y^{(k)} \leq t^{(k)}
\end{cases}.
\end{equation}
Ultimately, the ground-truth failure rate $\hat{P}_f $ is defined as:
\begin{equation}
\hat{P}_f = \int_\mathcal{X} I\left( \x \right) p(\x) d\x.
\end{equation}
}

\vspace{-0.1in}
\subsection{Gaussian process}
\vspace{-0.05in}
Gaussian process (GP) is a common choice as a surrogate model for building the input-output mapping for complex computer code due to its flexibility and uncertainty quantification.
We can approximate the black-box function $f_0(\x)$ by placing a GP prior:
$f_0({\x})| \pmb{\theta}\sim {\mathcal{GP}}(m({\x}),k({\x}, {\x}'|\pmb{\theta})),$

where the mean function is normally assumed zero, i.e.,  $m_0({\x})\equiv 0$, by virtue of centering the data. 
The covariance function can take many forms, the most common is the automatic relevance determinant (ARD) kernel
$    k({\x}, {\x}'|\pmb{\theta})=\theta_0\exp(-({\x}-{\x}')^T diag (\theta_1,\hdots,\theta_l) ({\x}-{\x}'))$, with $\pmb{\theta}$ denoting the hyperparameters.
Other kernels like Rational Quadratic (RQ), Periodic (PERD), and Matern can also be used depending on the application.

For any given design variable \({\x}\), \(f({\x})\) is now considered a random variable, and multiple observations  \(f({\x}_i)\) form a joint Gaussian with covariance matrix \(\K=[K_{ij}]\) where \(K_{ij}=k({\x}_i, {\x}_j)\). With a consideration of the model inadequacy and numerical noise \(\varepsilon \sim \mathcal{N}(0,\sigma^2)\), the likelihood $\Lcal$ is:
\begin{equation}
  \label{eq: GP likelihood}
  \Lcal=-\frac{1}{2}{\y}^T  (\textbf{K} +\sigma^2\I) ^{-1} {\y} -\frac{1}{2}\ln|\textbf{K} + \sigma^2\I|  - \frac{1}{2}\log(2\pi).
\end{equation}
The hyperparameters \(\pmb{\theta}\) in kernels are estimated via maximum likelihood estimates (MLE) of $\Lcal$.

Conditioning on \(\y\) gives a predictor posterior $\hat{f}(\x)\sim\mathcal{N}(\mu ({\x}), v({\x} ))$ 
\begin{equation}\label{postpredA}
  \small
  \mu ({\x} ) = {\k}^T({\x}) \left( \textbf{K} + \sigma^2 \I \right)^{-1}\y;
  v ({\x} ) = k({\x}, {\x} ) -{\k}^T({\x}) \left( \textbf{K} + \sigma^2 \I \right)^{-1} {\k}({\x})
\end{equation}

\cmt{
For any fixed ${\x}$, $f({\x})$ is a random variable. A collection of values $f({\x}_i)$, $i=1,\hdots,N$, on the other hand, is a partial realization of the GP. Realizations of the GP are deterministic functions of ${\x}$. 
The main property of GPs is that the joint distribution of $f({\x}_i)$, $i=1,\hdots,N$, is multivariate Gaussian.
Assuming the model inadequacy $\varepsilon \sim \mathcal{N}(0,\sigma^2)$ is also a Gaussian, with the prior and available data $\y=(y_1,\hdots,y_N)^T$, we can derive the model likelihood
\begin{equation}
    ({\y}^T  (\textbf{K} +\sigma^2\I) ^{-1} {\y} -\ln|\textbf{K} + \sigma^2\I|  - \log(2\pi) )/2
\end{equation}

where the covariance matrix $\K=[K_{ij}]$, in which $K_{ij}=k({\x}_i, {\x}_j)$, $i,j=1,\hdots,N$.
The hyperparameters $\pmb{\theta}$ are normally obtained from point estimates~\cite{rasmussen2006gaussian} by maximum likelihood estimate (MLE) of w.r.t. $\btheta$.
The joint distribution of $\y$ and $f(\x)$ also form a joint Gaussian distribution.

Conditioning on ${\y}$ provides the conditional Gaussian distribution at ${\x}$~\cite{rasmussen2006gaussian} with mean and variance $  \hat{f}({\x})|{\y},\pmb{\theta}\sim \mathcal{N}\left(\mu ({\x}|\pmb{\theta}), v ({\x},{\x}'|\pmb{\theta})\right),$ where

\begin{equation}\label{postpredA}
\begin{aligned}
\mu ({\x} ) &= {\k}({\x})^T \left( \textbf{K} + \sigma^2 \I \right)^{-1}\y\\ 
v ({\x} ) &=  \sigma^2 + k({\x}, {\x} ) -{\k}^T({\x}) \left( \textbf{K} + \sigma^2 \I \right)^{-1} {\k}({\x})
\end{aligned}
\end{equation}
}

\vspace{-0.2in}
\subsection{Bayesian Optimization}
\vspace{-0.05in}
To maximize $f(\x)$, we can optimize $\x$ by sequentially quarrying points such that each point shows an improvement $I(\x) = \max( \hat{f}(\x) - y^\dagger, 0)$, where $y^\dagger$ is the current optimal and $\hat{f}(\x)$ is the predictive posterior in \Eqref{postpredA}.
The possibility for $\x$ giving improvements is 
\begin{equation}
  \label{eq: PI}
  PI(\x) = \Phi\left( {(\mu(\x) - y^\dagger)}/ {\sigma(\x)}\right),
\end{equation}
which is the probability of improvement (PI).
For a more informative solution, we can take the expectative improvement (EI) over the predictive posterior:
\begin{equation}
    EI(\x) = (\mu(\x) - y^\dagger) \psi \left( u(\x) \right) + v(\x) \phi \left( u(\x) \right),
\end{equation}
where $\psi(\cdot)$ and $\phi(\cdot)$ are the probabilistic density function (PDF) and cumulative density function (CDF) of standard normal distribution, respectively.

The candidate for the next iteration is selected by $ \argmax_{\x \in \mathcal{X}} EI(\x)$ with gradient descent methods, e.g., L-BFGS-B.
Rather than looking into the expected improvement, we can approach the optimal by exploring the areas with higher uncertainty, a.k.a, upper confidence bound (UCB)
\begin{equation}
  UCB(\x)= \left(\mu(\x) + \beta v(\x) \right),
\end{equation}

where $\beta$ controls the tradeoff between exploration and exploitation. 

\begin{figure}
  \vspace{-0.2in}
  \centering
  \begin{subfigure}[c]{0.47\linewidth}
    \caption{\small Neural Kernel}
       \label{1}
    \includegraphics[width=\linewidth]{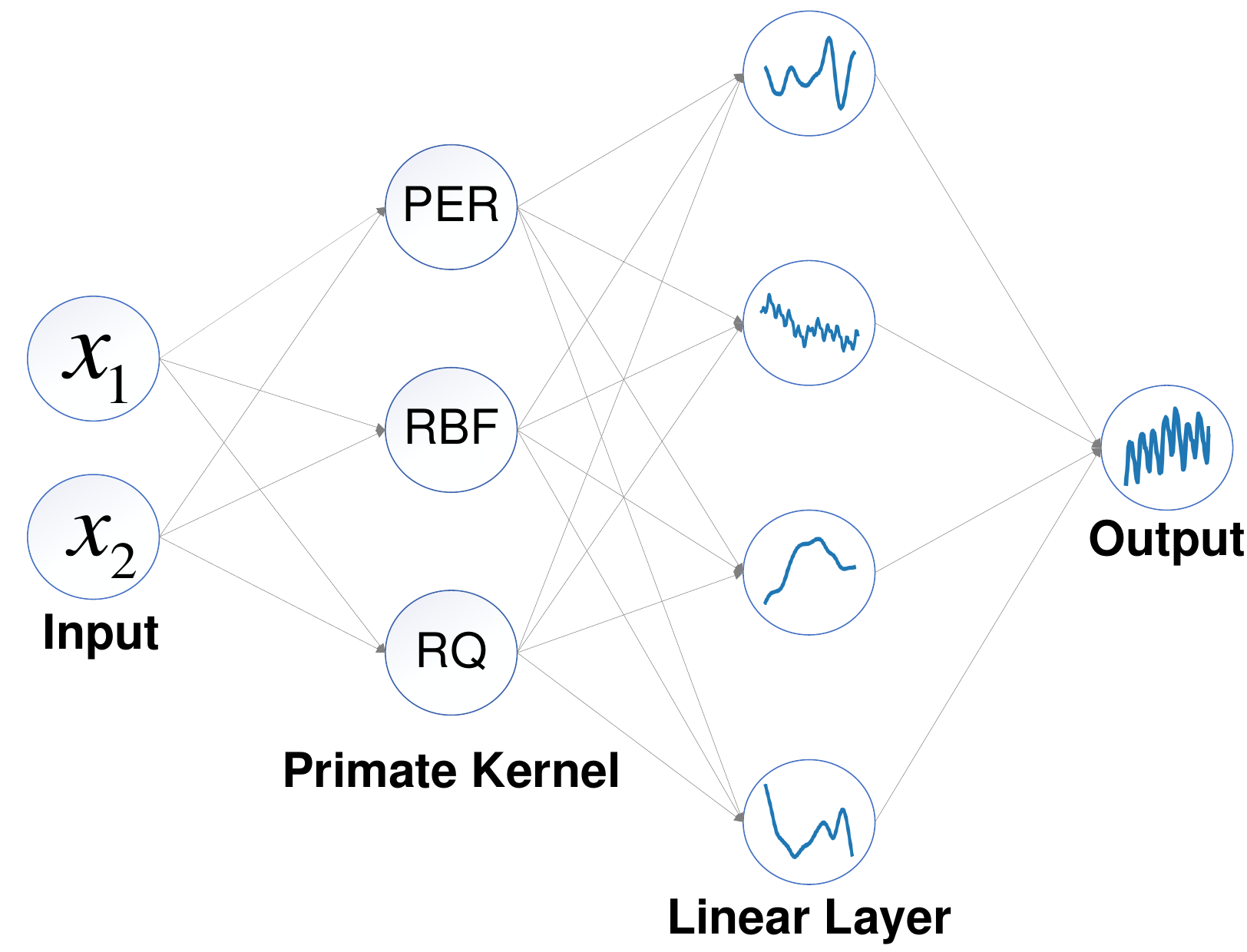}
  \end{subfigure}
  \begin{subfigure}[c]{0.42\linewidth}
    \caption{\small Assessment}
    \label{2}
      \includegraphics[width=\linewidth]{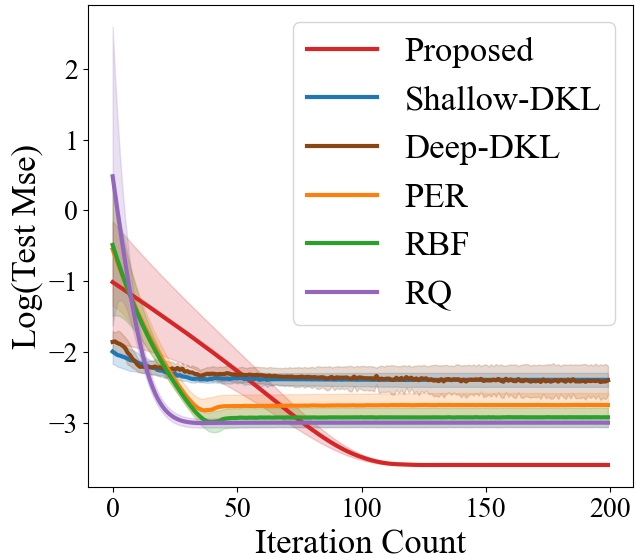}
  \end{subfigure}
  \vspace{-0.15in}
  \caption{Neural kernel and assessments}
  \label{fig: Neural Kernel}
  \vspace{-0.2in}
  \end{figure}
\vspace{-0.1in}
\section{Proposed methodologies}
\vspace{-0.05in}
\subsection{Automatic Kernel Learning: Neural Kernel}
\vspace{-0.05in}
Choosing the right kernel function is vital in BO. Despite DKL's success, creating an effective network structure remains challenging.
Following \cite{sun2018differentiable}, we introduce Neural Kernel (Neuk) as a basic unit to construct an automatic kernel function.

Neuk is inspired by the fact that kernel functions can be safely composed by adding and multiplying different kernels.
This compositional flexibility is mirrored in the architecture of neural networks, specifically within the linear layers. Neuk leverages this concept by substituting traditional nonlinear activation layers with kernel functions, facilitating an automatic kernel construction.

To illustrate, consider two input vectors, $\x_1$ and $\x_2$. In Neuk, these vectors are processed through multiple kernels $\{h_i(\x,\x')\}_i^{N_k}$, each undergoing a linear transformation as follows:
\begin{equation}
  h_i(\x_1,\x_2)={h_i(\W^{(i)}\x_1+\b^{(i)},\W^{(i)}\x_2+\b^{(i)})}
\end{equation}
Here, $\W^{(i)}$ and $\b^{(i)}$ represent the weight and bias of the $i$-th kernel function, respectively, and $h_i(\cdot)$ is the corresponding kernel function.
Subsequently, latent variables $\z$ are generated through a linear combination of these kernels:
\begin{equation}
  \z = \W^{(z)} \h(\x_1,\x_2) + \b^{(z)}, 
\end{equation}
where $\W^{(z)}$ and $\b^{(z)}$ are the weight and bias of the linear layer, and $\h(\x_1,\x_2) = [h_1(\x_1,\x_2),\hdots,h_{d_l}(\x_1,\x_2)]^T$.
This configuration constitutes the core unit of Neuk. For broader applications, multiple Neuk units can be stacked horizontally to form a Deep Neuk (DNeuk) or vertically for a Wide Neuk (WNeuk). However, in this study, we utilize a single Neuk unit, finding it sufficiently flexible and efficient without excessively increasing the parameter count.

The final step in the Neuk process involves a nonlinear transformation applied to the latent variables $\z$, ensuring the semi-positive definiteness of the kernel function,
\begin{equation}
  k_{neuk}(\x_1,\x_2) = \exp( \textstyle \sum z_j + b^{(k)}).
\end{equation}
A graphical representation of the Neuk architecture is shown in \Figref{fig: Neural Kernel} along with small experiments demonstrating its effectiveness in predicting performance in a 180nm second-stage amplification circuit (see Section \ref{sec: exp}) with 100 training and 50 testing data points.

\cmt{
Designing an appropriate model for surrogate-based optimization poses significant challenges, especially in GP where the selection of the kernel function is crucial. 
Despite the success of DKL, we found that designing a proper network is equally challenging (as shown in \Figref{fig: Neural Kernel} \Wei{add here}).
To address this challenge, we borrow the concept of neural network and use Neural Kernel (NK) \cite{NeuralKernel}.
NE utilizes the fact that a new kernel function can be safely composed (ensuring that the resulting kernel is positive definite) by adding and multiplying operation, which is contained in a neural network's linear layer.
Thus, the NK is a neural network kernel function that performs linear and nonlinear transformations on the base kernel.
More specifically, the NK takes two input vectors, $x_{1}$ and $x_{2}$, and outputs a kernel function value, denoted as $k\left (x_{1},  x_{2} \right)$.
In the first layer of NK, the original inputs are linearly transformed before they are fed into the kernel layer, which consists of multiple base kernel functions ( including linear, RBF (Radial Basis Function), and RQ (Rational Quadratic) kernels).
These results are then linearly combined to generate latent variables $\z$, which capture stationary and non-stationary features of the data.
Finally, $\z$ undergoes a nonlinear transformation to an exponential operation (to ensure that the correlation is always semi-positive definite) to generate the final correlation. We call this a basic neural kernel unit (NKU).
}

\cmt{
The Neural Kernel is a specialized kernel function based on neural networks. It is designed to perform both linear and nonlinear transformations on the foundational kernel. This function processes two input vectors, $x_{1}$ and $x_{2}$, and produces a kernel function value, symbolized as $k\left (x_{1},  x_{2} \right)$.

Diving into the architecture of the Neural Kernel, the first layer integrates various foundational kernel functions such as linear, RBF (Radial Basis Function), and RQ (Rational Quadratic) kernels. 
These functions, fundamental to our experiments, possess individual hyperparameters like length scale and amplitude, which are subsequently numerically optimized using gradient backpropagation.

Following the foundational layer is a linear one, which allocates specific weights and biases to each kernel function, facilitating potential adjustments in optimization phases that follow.

Concluding the architecture is the nonlinear layer, placed post the linear layer. This layer applies a nonlinear transformation to the outcomes of the linear combinations. The said transformation is mathematically articulated as $o=f\left (z  \right )$, where $z$ denotes the output of its preceding layer. In this setup, the nonlinear activation function is chosen as \( f(z) = e^z \), which ensures the continuity and closure of the kernel function.
}


\begin{figure}[t]
  \vspace{-0.1in}
    \centering
    \includegraphics[width=1\linewidth]{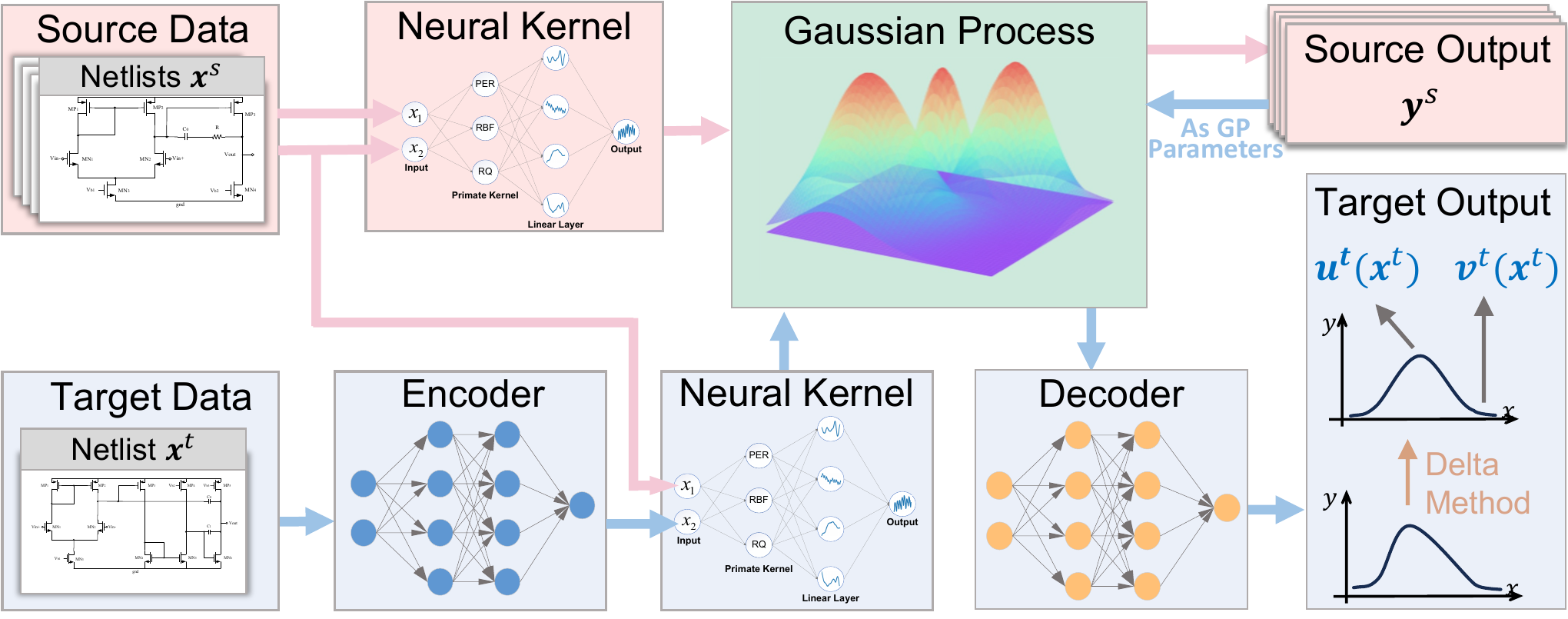}
    \vspace{-0.28in}
    \caption{Knowledge Alignment and Transfer (KAT)-GP}
     \label{Encoder-decoder GP}
     \vspace{-0.3in}
 \end{figure}
\vspace{-0.1in}
\subsection{Knowledge Alignment and Transfer}
\vspace{-0.05in}
In the literature, transfer learning is predominantly based on deep learning, wherein the knowledge is encoded within the neural network weights, facilitating transfer learning through the method of fine-tuning these weights on a target dataset.
Contrastingly, GPs present a fundamentally different paradigm. The predictive capability of GPs is intrinsically tied to the source data they are trained on (see \Eqref{postpredA}). This reliance on data for prediction underscores a significant challenge in applying transfer learning to GPs.

To address the challenge of applying transfer learning to GPs, we propose an innovative encoder-decoder structure, which we refer to as Knowledge Alignment and Transfer (KAT) in GPs. This approach retains the intrinsic knowledge of the source GP while aligning it with the target domain through an encoder and decoder mechanism.

Consider a source dataset $\Dcal^{(s)}=\{(\x^{(s)}_i, \y^{(s)}_i)\}$, on which the GP model $GP(\x)$ is trained, and a target dataset $\Dcal^{(t)}=\{(\x^{(t)}_i, \y^{(t)}_i)\}$. The first step involves introducing an encoder $\E(\x)$, which maps the target input $\x^{(t)}$ into the source input space $\x^{(s)}$. This encoder accounts for potential differences in dimensionality between the source and target datasets, effectively managing any compression or redundancy.

The target outputs may have different value ranges or even quantities from the source outputs. We employ a decoder $\D(\y^{(s)})$ that transforms the output of the GP $G(\x)$ to match the target output $\y^{(t)}$. Thus, the KAT-GP for the target domain is expressed as $\y^{(s)}=\D(GP(\E(\x^{(t)})))$.

A key aspect of this approach is that the original observations in the source GP are preserved, while the encoder and decoder are specifically trained to align and transfer knowledge between the source and target domains. The encoder and decoder themselves can be complex functions, such as deep neural networks, and their specific architecture depends on the problem and available data.
In this work, both the encoder and decoder are small shallow neural networks with linear($d_{in}\times {32}$)-sigmoid-linear(${32}\times d_{out}$) structure, where $d_{in}$ and $d_{out}$ are the input and output dimension during implementation.

It is important to note that unless the decoder is a linear operator, KAT-GP is no longer a GP and does not admit a closed-form solution. We approximate the predictive mean and variance of the overall model through the Delta method, which employs Taylor series expansions to estimate the predictive mean $\bmu^{(t)}(\x^{(t)})$ and covariance $\S^{(t)}(\x^{(t)})$ of the transformed output:
\begin{equation}
    \begin{aligned}
        \mu^{(t)}(\x^{(t)}) &= D\left( \bmu^{(s)}(E(\x^{(t)})) \right); \quad
        \S^{(t)}(\x^{(t)}) &= \J \S^{(s)} \J^T,
    \end{aligned}
\end{equation}
where $\bmu^{(s)}(\x^{(s)})$ and $\S^{(s)}(\x^{(s)})$ are the predictive mean and covariance of $GP(\x)$, respectively, and $\J$ is the Jacobian matrix of $\D(\mu^{(t)}(\x^{(t)}) )$ with respect to $\bmu^{(t)}(\x^{(t)}) $.
Thus, training of KAT-GP involves maximizing the log-likelihood in \Eqref{eq: KAT-GP ll} using gradient descent w.r.t the parameters of the encoder and decoder, as well as the hyperparameters of the neural kernel.
\begin{equation}
  \label{eq: KAT-GP ll}
    \mathcal{L} = \sum \log \mathcal{N}(\y^{(t)}_i|D(\bmu(E(\x^{(t)}_i))), \J_i \S_i \J_i^T + \sigma^2_t \I,).
\end{equation}
KAT-GP is illustrated in \Figref{Encoder-decoder GP}. It offers the first knowledge transfer solution for GP with different design and performance space.


\cmt{
---
To resolve this challenge, we introduce an encoder-decoder structure to enable knowledge alignment and transfer (KAT) in GPs.
Denote the source dataset as $\Dcal^{(s)}=\{(\x^{(s)}_i, \y^{(s)}_i)\}$, based on which $GP(\x)$ is trained, and the target dataset as $\Dcal^{(t)}=\{(\x^{(t)}_i, \y^{(t)}_i)\}$. 
we first introduce a Encoder $\E(\x)$ to encode the target input $\x^{(t)}$ to the source input $\x^{(s)}$. Note that the dimensionality can be different between the source and target datasets and the compression/redundancy is handled by the encoder.

To transfer the output of the source domain to the target domain, we decode the output of the GP model $G(\x)$ to the target output $\y^{(t)}$ by a decoder $\D(\y^{(s)})$. 
Thus, the KAT-GP for target domain is $\y^{(s)}=\D(GP(\E(\x^{(t)})))$.
The key ingredient is that the knowledge (\ie observations) in the source GP is preserved; the encoder and decoder are trained to align the source and target knowledge.
The encoder and decoder can be any complex functions, e.g., deep neural networks. This will depend on the specific problem and the available data.
In this work, both the encoder and decoder are small shallow neural networks with linear($d_{in}\times {32}$)-sigmoid-linear(${32}\times d_{out}$) structure, where $d_{in}$ and $d_{out}$ are the input and output dimension during implementation.

Unless the decoder is a linear operator, KAT-GP is no longer a GP and does not admit a closed-form solution.
However, we can use the posterior mean and variance in \Eqref{postpredA} to approximate the predictive mean and variance of the overall model.
More specifically, we can use the Delta method to approximate the predictive mean $\bmu^{(t)}(\x^{(t)})$ and covariance $\S^{(t)}(\x^{(t)})$ of a transformation of the output using Taylor series expansions:

\begin{equation}
    \begin{aligned}
        \mu^{(t)}(\x^{(t)}) &= D\left( \bmu^{(s)}(E(\x^{(t)})) \right); \quad
        \S^{(t)}(\x^{(t)}) &= \J \S^{(s)} \J^T,
    \end{aligned}
\end{equation}
 $\bmu^{(s)}(\x^{(s)})$ and $\S^{(s)}(\x^{(s)})$ are the predictive mean and covariance of $GP(\x)$, respectively, and $\J$ is the Jacobian matrix of $\D(\mu^{(t)}(\x^{(t)}) )$ w.r.t. $\bmu^{(t)}(\x^{(t)}) $.

Thus, training of the target model can be conducted using MLE for the log-likelihood 
\begin{equation}
  \label{eq: KAT-GP ll}
    \mathcal{L} = \sum \log \mathcal{N}(\y^{(t)}_i|D(\bmu(E(\x^{(t)}_i))), \J_i \S_i \J_i^T + \sigma^2_t \I,),
\end{equation}
where we introduce $\sigma^2_t$ for the noise variance in the target dataset. MLE is conducted using gradient-based optimization, e.g., L-BFGS-B, w.r.t the parameters of the encoder and decoder, and the hyperparameter in the neural kernel, whose derivative is available.
A summary of the KAT-GP is shown in \Figref{Encoder-decoder GP}.
}

\cmt{
\subsection{Transfer Learning: Encoder-decoder GP}  
In the existing body of research, transfer learning is predominantly based on deep learning methodologies, wherein the accumulated knowledge is embedded within the neural network weights. This allows for relatively straightforward adaptation through the fine-tuning of these weights on a target dataset. However, the inherent nature of Gaussian Processes (GPs) differs substantially in that they rely explicitly on data for their predictive power. This characteristic is exemplified in the predictive mean as outlined in Equation \ref{postpredA}, which represents predictions as a linear combination of observed data. Consequently, applying transfer learning to GPs presents a more complex challenge compared to deep learning models.

To address this challenge, we propose a novel encoder-decoder architecture designed to facilitate knowledge alignment and transfer (KAT) within the framework of GPs. Let the source dataset be defined as \(\mathcal{D}^{(s)} = \{(\mathbf{x}^{(s)}_i, y^{(s)}_i)\}_{i=1}^{N^{(s)}}\), from which a GP model, denoted as \(G(\mathbf{x})\), is trained. Correspondingly, the target dataset is denoted as \(\mathcal{D}^{(t)} = \{(\mathbf{x}^{(t)}_i, y^{(t)}_i)\}_{i=1}^{N^{(t)}}\). We introduce an encoder, \(E(\mathbf{x})\), that maps the target inputs, \(\mathbf{x}^{(t)}\), to a space aligned with the source inputs, \(\mathbf{x}^{(s)}\). The encoder begins as an identity function with an added perturbation, allowing for an initial baseline transformation. The source domain GP outputs are then adapted to the target domain outputs via a decoder, \(D(y^{(s)})\), forming the comprehensive model \(y^{(s)} = D(G(E(\mathbf{x}^{(t)})))\). This structure ensures the preservation of source GP knowledge while permitting the alignment of source and target domains through encoder and decoder training.

The encoder and decoder may be represented by any suitably complex function, such as deep neural networks, with the specific choice dictated by the problem context and data availability. In this study, the encoder and decoder are instantiated as small neural networks employing a linear-sigmoid-linear structure, consisting of 2-10-1 neurons \stutodo{change here}. It should be noted that unless the decoder is a purely linear operation, the resultant model no longer retains the closed-form solution property intrinsic to GPs.

Despite this, we can leverage the posterior mean and variance from \Eqref{postpredA} to estimate the predictive mean and variance of the integrated model. More precisely, the Delta method facilitates the approximation of the mean and variance for the transformed outputs. The mean of \(\y^{(t)}\) aligns with the mean of \(G(\mathbf{x})\), denoted as \(\mu^{(s)}(\mathbf{x}) = \mu(\mathbf{x})\). The covariance of \(y^{(s)}\) is expressed as \(\mathbf{J} \mathbf{\Sigma} \mathbf{J}^T\), where \(\mathbf{J}\) is the Jacobian matrix of \(D(y^{(s)})\) with respect to \(y^{(s)}\), and \(\mathbf{\Sigma}\) represents the predictive covariance matrix of \(G(\mathbf{x})\), i.e., \(\mathbf{\Sigma} = \mathbf{K} + \sigma^2 \mathbf{I}\).

Training of the target model thus proceeds via maximum likelihood estimation (MLE) for the log-likelihood, which can be formalized as:
\begin{equation}
  \mathcal{L} = \sum_{i=1}^{N^{(t)}} \log \mathcal{N}(\y^{(t)}_i|D(\bmu(E(\x^{(t)}_i))), \J_i \S_i \J_i^T).
\end{equation}
This approach not only ensures a theoretically grounded transfer of knowledge between datasets but also provides a practical framework for the application of Gaussian Processes in scenarios where transfer learning is essential.
}

\vspace{-0.1in}
\subsection{Modified Constrained MACE}
\vspace{-0.05in}
In transistor sizing, it is crucial to harness the power of parallel computing by running multiple simulations simultaneously.
One of the most popular solutions, MACE \cite{zhang2021efficient} resolves this challenge by proposing candidates lying on the Pareto frontier of objectives \begin{small}
    $\{{UCB}(\x), {PI}(\x), {EI}(\x), {PF}(\x),
\sum_{i=1}^{N_c} \max(0, u_i(\x)), \sum_{i=1}^{N_c} \max(0, \frac{u_i(\x)}{v_i(\x)})\}$\end{small}\\
using Genetic searching NSGA-II.
Here, ${PF}(\x)$ is the probability of feasibility, which use \Eqref{eq: PI} with all constraint metrics, \ie ${PF}(\x)=\prod_i^{N_c} \Phi({u_i(\x)-C_i}/{v_i(\x)})$.

Despite its success, MACE suffers from high computational complexity, as it requires a Pareto front search with six correlated objectives. To mitigate this issue, we consider the constraint as an additional objective for the primal metric $f_0(\x)$ and the multi-objective optimization becomes 
\vspace{-0.02in}
\begin{equation}
  \argmax \{{UCB}(\x), {PI}(\x) , {EI}(\x)  \} \times PF(\x).
\end{equation}
This reduction in dimensionality significantly improves efficiency, as the complexity grows exponentially with the number of objectives while maintaining the same level of performance. Empirically, we do not observe any performance degradation.

\vspace{-0.1in}
\subsection{Selective Transfer Learning with BO}  
\vspace{-0.05in}
While transfer learning proves effective in numerous scenarios, its utility is not universal, particularly when the source and target domains differ significantly, such as between an SRAM and an ADC. It's important to note that even when the source and target datasets have an equal number of points, the utility of KAT-GP may not be immediately apparent. However, our empirical studies reveal that source and target data often exhibit distinctly different distributions, with varying concentration regions. This divergence can provide valuable insights for the optimization of the target circuit, even when the target data exceeds the source data in size. The effectiveness of transfer learning in such cases is inherently problem-dependent, requiring adaptable strategies for diverse scenarios.

To address these challenges, we propose a Selective Transfer Learning (STL) strategy, which synergizes with the batch nature of the MACE algorithm to optimize the benefits of transfer learning. This approach involves training both a KAT-GP model and a GP model (referred to as NeukGP, equipped with a Neural Kernel) exclusively on the target data. 
%
During the Bayesian Optimization (BO) process, each model collaborates with MACE to generate proposal Pareto front sets, denoted as $\mathcal{P}i$ (with $i=1,2$ in this context). 
We randomly select $\frac{w{1}}{w{1}+w_2} N_B$ points from $\mathcal{P}1$ to form $\mathcal{A}1$, and $\frac{w{2}}{w{1}+w_2} N_B$ points from $\mathcal{P}_2$ to form $\mathcal{A}_2$.
Points in $\mathcal{A}1$ and $\mathcal{A}2$ are then simulated and evaluated.
The weights are initialized with the number of samples and updated based on the number of simulations that improve the current best, \ie
\vspace{-0.05in}
\begin{equation}
  \label{eq: stl}
w_i = w_i + |f(\mathcal{A}i) > y^{\dagger}|_n,
\end{equation}
where $|f(\mathcal{A}i) > y^{\dagger}|_n$ represents the number of points in $\mathcal{A}i$ that surpass the current best objective value $y^{\dagger}$, and $f$ can be the constrained objective or the Figure of Merit (FOM). The STL algorithm is summarized in \Algref{algo2}.

\cmt{
---
Despite that transfer learning is effective in many cases, it is not always beneficial, which is intuitive in our case if the source and target circuits are very different, \eg, an SRAM and an ADC.
It is worth noting that when the source and target data contain the same number of points, it seems unnecessary to use KAT-GP at first glance.
However, through our empirical study, the source and target data can have quite different distributions, \ie different concentration regions, which offers useful information for optimization of the target circuit even when the target data is larger than the source data.
Certainly, these are problems dependent and a flexible way must be developed to adapt to different scenarios.

To address this problem, we propose a selective transfer learning (STL) approach, which takes advantage of the batch nature of MACE to maximize the benefit of transfer learning.
More specifically, except for training a KAT-GP model, we also train a GP model directly on the target data, which we call NeukGP (with the Neural Kernel).
Each GP is given an initial weight.
During the BO, each model is combined with MACE to generate a proposal Pareto front sets $\mathcal{P}_i$ (in this case $i=1,2$).
Based on the weights and the batch size $N_B$, we randomly select $ \frac{w_{1}}{w_{1}+w_2} N_B$ points from $\mathcal{P}_1$ and $\frac{w_{2}}{w_{1}+w_2} N_B$ points from $\mathcal{P}_2$ to form action set $\mathcal{A}_1$ and $\mathcal{A}_2$, respectively.
The points in $\mathcal{A}_1$ and $\mathcal{A}_2$ are then simulated and evaluated.
The weights are then updated based on the number of simulations that improve the current best, \ie
\vspace{-0.1in}
\begin{equation}
  \label{eq: stl}
  w_i = w_i + |f(\mathcal{A}_i) > f_{best}|_n,
\end{equation} 
where $|f(\mathcal{A}_i) > f_{best}|_n$ is the number of points in $\mathcal{A}_i$ that improve the current best, $f_{best}$, and $f$ is can either be the constrained objective or FOM.
The algorithm is summarized in \Algref{algo2}.
}

\vspace{-0.15in}
 \begin{algorithm}
   \caption{\ours with Selective Transfer Learning}
      \begin{algorithmic}[1]  \label{algo2}
      \REQUIRE Source dataset $\mathcal{D}_{s}$ , initial target circuit data $\mathcal{D}_{t}$, \# iterations $N_{I}$. 
      $B$, batch size $N_B$ per iteration \\
     \STATE Train KAT-GP on $\mathcal{D}_{s}$ and $\mathcal{D}_{t}$
     \STATE Train NeukGP based on $\mathcal{D}_{t}$
      \FOR{$i=1 \to N_{I} $} 
      \STATE Update KAT-GP and NeukGP based on $\mathcal{D}_{t}$
      \STATE Apply MACE to KAT-GP and NeukGP to generate proposal set $\mathcal{P}_{1}$ and $\mathcal{P}_{2}$.
      \STATE Form action set $\mathcal{A}_1$ and $\mathcal{A}_2$ by randomly selecting $ \frac{w_{1}}{w_{1}+w_2} N_B$ points from $\mathcal{P}_{1}$ and $\frac{w_{2}}{w_{1}+w_{2}} N_B$ points from $\mathcal{P}_{2}$.
      \STATE Simulate $\mathcal{A}_1$ and $\mathcal{A}_2$ and update $\mathcal{D}_{t}\leftarrow \mathcal{D}_{t} \cup \mathcal{A}_1 \cup \mathcal{A}_2 $.
      \STATE Update $w_{1}$ and $w_{2}$ based on \Eqref{eq: stl} and best design $\x_*$.
      \ENDFOR
    \end{algorithmic} 
  \end{algorithm}
  \vspace{-0.25in}


\cmt{
In constrained optimization contexts, we introduce an auxiliary constraint function, \(FI\), to increase the likelihood of selecting points that meet the constraints. This function combines the logarithms of the probabilities that each metric complies with its constraints:

\begin{equation}
\sum\limits_{i=1}^{n}{\log \{PI[{{\mu }_{i}}(x)-con_{i},{{\sigma }_{i}}(x)]\}}
\end{equation}

Here, \({{\mu }_{i}}(x)\) and \({{\sigma }_{i}}(x)\) represent the expected and variance values of various constraint metrics, with \(con_{i}\) indicating the constraint values for different metrics and \(n\) the number of constraints.

The optimization process employs the NAGSII algorithm, identifying optimal points along the Pareto frontier. A random selection from these Pareto-optimal points is used for simulation predictions, forming the basis for further refinement of the surrogate model. The algorithmic steps are outlined in Algorithm 1.
}

\cmt{
\subsection{Porfolio Acquisition Strategy With constraint}
Selecting the ideal acquisition function is pivotal to the efficacy of any optimization process. To harness the unique advantages of various acquisition functions, we have devised a multi-objective acquisition function rooted in the MACE framework. This strategy aims to simultaneously maximize three distinct acquisition functions: LCB, PI, and EI, to discern the optimal set of points residing on the Pareto frontier. Formally, the objective can be represented as:
\begin{equation}
\text{maximize } -\mathrm{LCB}(x),-\mathrm{PI}(x),-\mathrm{EI}(x)
\end{equation}
To adapt this to scenarios requiring objective value maximization, the negative signs preceding the acquisition functions are simply omitted.

For problems encompassing constraints, we introduce an auxiliary constraint function, denoted as $FI$. This function augments the probability that the chosen set of points satisfies the imposed constraints. $FI$ is articulated as the aggregate of the logarithms of the probabilities ensuring each metric adheres to its respective constraints:
\begin{equation}
\sum\limits_{i=1}^{n}{\log \{PI[{{\mu }_{i}}(x)-co{{n}_{i}},{{\sigma }_{i}}(x)]\}}
\end{equation}
In this expression, ${{\mu }_{i}}(x)$ and ${{\sigma }_{i}}(x)$ denote the anticipated and variance values of alternative constraint metrics, respectively. Meanwhile, $co{{n}_{i}}$ symbolizes the constraint values associated with diverse indicators, and $n$ quantifies the constraint conditions.

Having delineated the multi-objective functions as detailed above, optimization is undertaken employing the NAGSII algorithm. This culminates in the identification of optimal points dispersed along the Pareto frontier. Following this, a random assortment of points is chosen from this Pareto-optimal set for simulation prediction. These points serve as the foundation for subsequent surrogate model refinement. The sequence of the algorithm's operations is elucidated in Algorithm 1.
}

\def\Figref#1{Fig.~\ref{#1}}
\def\Tabref#1{Table~\ref{#1}}

\begin{figure*}
    \vspace{-0.2in}
    \centering
\begin{subfigure}[b]{0.27\linewidth}
    \caption{Two-stage operational amplifier}
    \label{a}
    \vspace{-0.05in}
    \includegraphics[width=\linewidth]{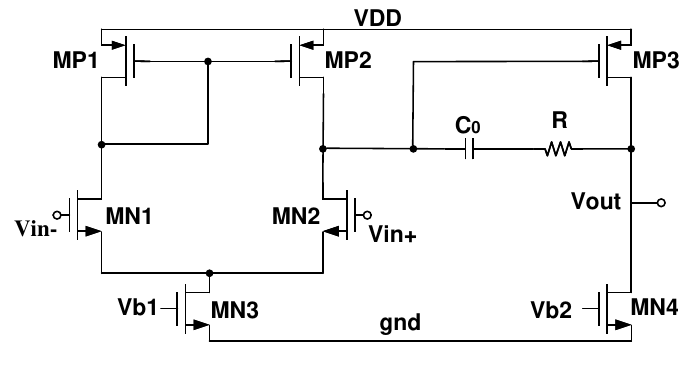}
\end{subfigure}
\begin{subfigure}[b]{0.32\linewidth}
    \caption{Three-stage operational amplifier}
    \label{b}
    \vspace{-0.05in}
    \includegraphics[width=\linewidth]{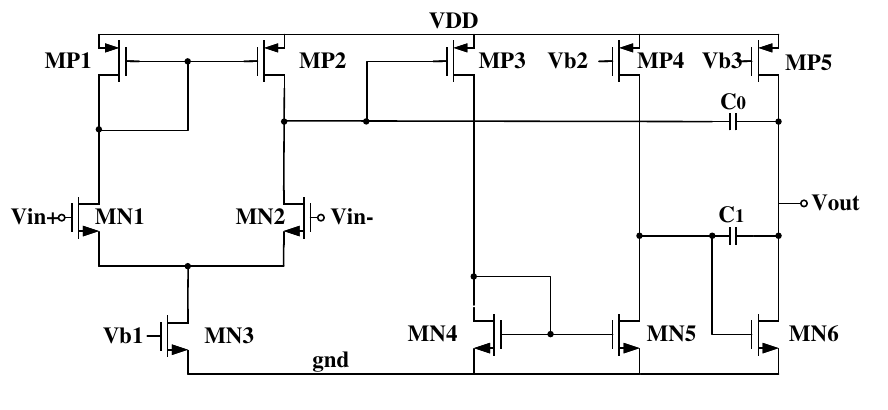}
\end{subfigure}
\begin{subfigure}[b]{0.31\linewidth}
    \caption{Bandgap}
    \label{c}
    \vspace{-0.05in}
    \includegraphics[width=\linewidth]{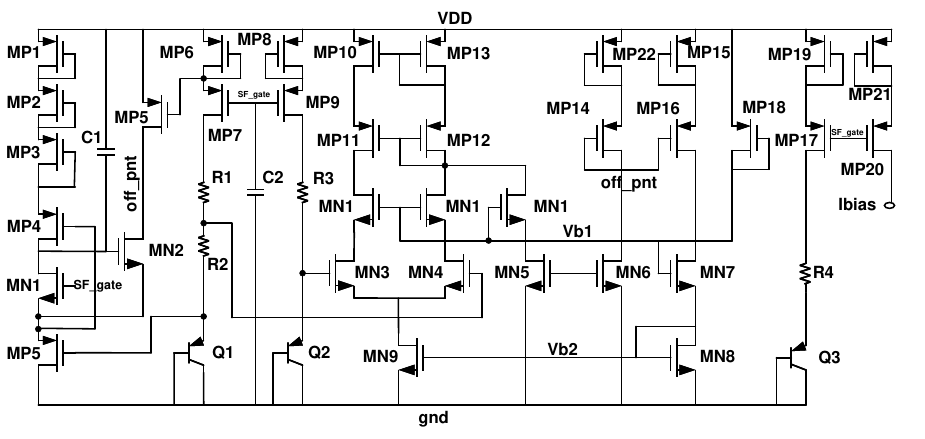}
\end{subfigure}
\vspace{-0.2in}
\caption{Schematic of the evaluation circuits}
\label{fig:circuits}
\vspace{-0.1in}
\end{figure*}

\vspace{-0.03in}
\section{Experiment}
\vspace{-0.05in}
\label{sec: exp}
To assess the effectiveness of \ours, we conducted experiments on three analog circuits: a Two-stage operational amplifier (OpAmp), a Three-stage OpAmp, and a Bandgap circuit (depicted in \Figref{fig:circuits}).

\noindent\textbf{Two-stage Operational Amplifier (OpAmp)} focuses on optimizing the length of the transistors in the first stage, the capacitance of the capacitors, the resistance of the resistors, and the bias currents for both stages. The objective is to minimize the total current consumption ($I_{total}$) while meeting specific performance criteria: phase margin (PM) greater than 60 degrees, gain-bandwidth product (GBW) over 4MHz, and a gain exceeding 60dB.
\begin{equation}
\argmin I_{total} ; \text{s.t.} ; PM > 60^\circ, GBW > 4MHz, Gain > 60dB.
\end{equation}
\noindent\textbf{Three-stage OpAmp} improves the gain beyond what a two-stage OpAmp can offer with an additional stage. This variant introduces more design variables, including the length of transistors in the first stage, the capacitance of two capacitors, and the bias currents for all three stages. The optimization is specified as:
\begin{equation}
\argmin I_{total} ; \text{s.t.} ; PM > 60^\circ, GBW > 2MHz, Gain > 80dB.
\end{equation}
\noindent\textbf{Bandgap Reference Circuit} is vital in maintaining precise and stable outputs in analog and mixed-signal systems-on-a-chip. The design variables include the length of the input transistor, the widths of the bias transistors for the operational amplifier, and the resistance of the resistors. 
The aim is to minimize the temperature coefficient (TC), with constraints on total current consumption ($I_{total}$ less than 6uA) and power supply rejection ratio (PSRR larger than 50dB):
\begin{equation}
\argmin TC ; \text{s.t.} ; I_{total} < 6uA, PSRR > 50dB.
\end{equation}
For the experiments, each method was repeated five times with different random seeds, and statistical results were reported.

Baselines were implemented with fine-tuned hyperparameters to ensure optimal performance. All circuits were implemented using 180nm and 40nm Process Design Kits (PDK). \ours is implemented in PyTorch with MACE\footnote{\href{https://github.com/Alaya-in-Matrix/MACE}{https://github.com/Alaya-in-Matrix/MACE}}.
Experiments were carried out on a workstation equipped with an AMD 7950x CPU and 64GB RAM.

\cmt{
---
To assess the proposed method, we conduct experiments on three circuits: Two-stage operational amplifier (OpAmp), three-stage OpAmp, and bandgap circuit, which are illustrated in \Figref{fig:circuits}.

The two-stage operational amplifier (OpAmp) uses the length of transistors of the first stage, the capacitance of the capacitors, the resistance of the resistors, and the bias currents for the first and second stages as design variables. 

We aim to minimize the total current consumption $I_{total}$ while satisfying the minimal requirements for phase margin (PM), gain-bandwidth product (GBW), and gain,

\begin{equation}
\argmin I_{total} \; s.t. \; PM > 60^\circ, GBW > 4MHz, Gain > 60dB.
\end{equation}

To improve the gain of a two-stage OpAmp, one can add an additional stage to form a three-stage OpAmp (shown in \Figref{b}), which is challenging with more design variables: the length of transistors of the first stage, the capacitance of the two capacitors, and the bias currents for the first, second and third stages. The optimization is

\begin{equation}
\argmin I_{total} \; s.t. \; PM > 60^\circ, GBW > 2MHz, Gain > 80dB.
\end{equation}

Bandgap reference circuits play a crucial role in analog and mixed-signal systems-on-a-chip, and their outputs must remain precise and stable to ensure the normal operation of the whole chip.

The design variable includes the length of the input transistor widths of the bias transistors for the operational amplifier and the resistance of the resistors

We aim to minimize the temperature coefficient (TC) while satisfying the constraint on the total current consumption $I_{total}$ and power supply rejection ratio (PSRR),
\begin{equation}
\argmin TC \; s.t. \; I_{total} < 6uA, PSRR < -50dB.
\end{equation}
For our experiment, all methods are repeated five times with different random seeds, and the statistical results are reported.
The baseline methods are implemented using our best effort to fine-tune the hyperparameters to achieve the best performance.

All circuits were implemented using 180nm and 40nm PDK.
All experiments are conducted on a workstation with an AMD 7950x CPU and 64GB RAM.

\href{https://github.com/automl/SMAC3}{https://github.com/automl/SMAC3}
\href{https://github.com/Alaya-in-Matrix/MACE}{https://github.com/Alaya-in-Matrix/MACE}
\href{https://github.com/pytorch/pytorch}{https://github.com/pytorch/pytorch}
\Wei{MACE address + pytoch address + SMAC address}
}

\cmt{
The two-stage OpAmp and the three-stage OpAmp are designed using 180nm and 40nm processes. The bandgap circuit is designed using a 180nm process. \Wei{is this right?}
The design variables in these circuits include the length of transistors, the capacitance of the capacitors, the resistance of the resistors, and the bias currents for the first and second stages. The performance metrics chosen for these circuits are gain, phase margin (PM), gain-bandwidth product (GBW), and power. 
For the experiment with a constrained BO algorithm, the design specifications are as follows:
}

\cmt{
\subsection{Two-stage operational amplifier}
\YYwrite{The circuit of the two-stage operational amplifier shown in Fig. \ref{1} is designed using 180nm and 40nm process. The 180nm is used to test the unconstrained and constrained BO algorithm, and transfer learning from 180nm to 40nm is performed to test the results of transfer learning. The design variables in this circuit include the length of transistors, the capacitance of the capacitors, the resistance of the resistors, and the bias currents for the first and second stages. . The performance metrics chosen for this circuit are gain, phase margin (PM), gain-bandwidth product (GBW), and power. For the experiment with a constrained BO algorithm, the design specifications are as follows: 
\begin{equation}
\label{eq3}
\begin{split}
Minimize\quad  I_{total}&        \\
    s.t.\quad\quad PM       &> 60°     \\       
               GBW      &> 4MHz   \\
               Gain     &> 60dB
\end{split}
\end{equation}
For the experiment with an unconstrained BO algorithm, we define the performance metrics as Figure of Merit (FOM), which is the weighted sum of normalized performance metrics as shown in Equation 1.

For this circuit, in the experiment with unconstrained BO algorithm, we compared the MACE algorithm with the SMAC and Random Search algorithms. We run the algorithms under the same conditions and perform 100 simulations to compare which algorithm yields the highest FOM among the points found. In the experiment involving knowledge transfer between technology nodes, we compare the performance of different algorithms under the prior knowledge of data from the 180nm two-stage amplification circuit. The goal is to determine which algorithm achieves the highest FOM among 100 fixed-cost simulations in the 40nm process. }

\subsection{Three-stage operational amplifier}
\YYwrite{The circuit of the Three-stage operational amplifier shown in Fig.\ref{2}  is designed using 180nm process. The circuit is used to test constrained and unconstrained BO algorithm. And we can also transfer the knowledge learned between the three-stage operational amplifier and the 180nm two-stage operational amplifier. For this circuit, the design variables include  the length of transistors, the capacitance of the capacitors $C_0$ and $C_1$, the resistance of the resistors, and the bias currents for the first, second and third stages. Similarly, the performance metrics for this circuit are gain, phase margin (PM), gain-bandwidth product (GBW), and power. For  the experiment with constrained BO algorithm, the design specifications are as follows: 
\begin{equation}
\label{eq2}
\begin{split}
Minimize\quad  i_{total}&        \\
    s.t.\quad\quad PM       &> 60°     \\      
               GBW      &> 2MHz   \\
               gain     &> 60dB
\end{split}
\end{equation}

For this circuit, in the experiment with \textcolor{red}{an} constrained BO algorithm, we compared the USeMOC algorithm with the MESMOC algorithm. Under the same conditions, involving 300 pre-sampled simulations and a fixed simulation budget of 500, we compare algorithms to determine which one finds the point with the minimum $i_{total}$ while satisfying the constraint conditions.
}

\subsection{Bandgap circuit}
\YYwrite{The schematic of the bandgap is shown in Fig.\ref{c}. The circuit is designed using 180nm process. For this circuit, the design variables include the lengths of transistors,  the widths of transistors and the resistance of the resistors. The performance metric for this circuit is temperature coefficient, power supply rejection ratio(PSRR) and power. This circuit is used to conduct the experiment with constrained BO algorithm.
For the experiment with constrained BO algorithm, the design specifications are as follows:
\begin{equation}
\label{eq2}
\begin{split}
Minimize\quad  TC&        \\
    s.t.\quad\quad i_{total}&< 6uA  \\
               PSRR     &< -50dB
\end{split}
\end{equation}

}
}

\begin{figure*}
    \vspace{-0.2in}
    \centering
    \begin{subfigure}[b]{0.29\linewidth}
    \caption{Two-stage OpAmp}
    \label{Two-stage OpAmp}
    \vspace{-0.04in}
    \includegraphics[width=\linewidth]{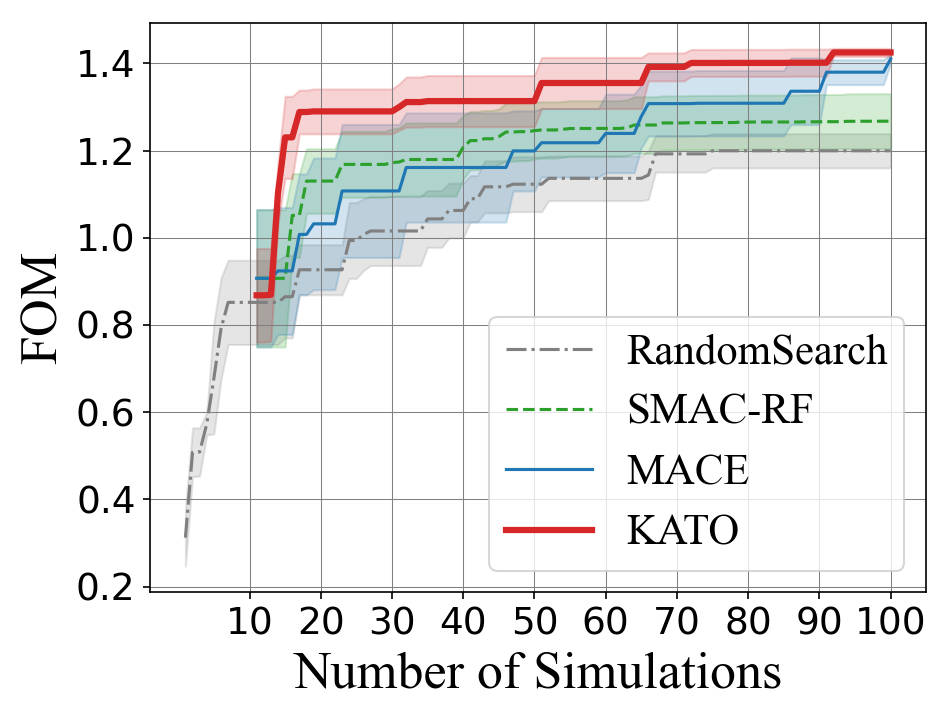}
\end{subfigure}
\begin{subfigure}[b]{0.29\linewidth}
    \caption{Three-stage OpAmp}
    \label{Three-stage OpAmp}
    \vspace{-0.04in}
    \includegraphics[width=\linewidth]{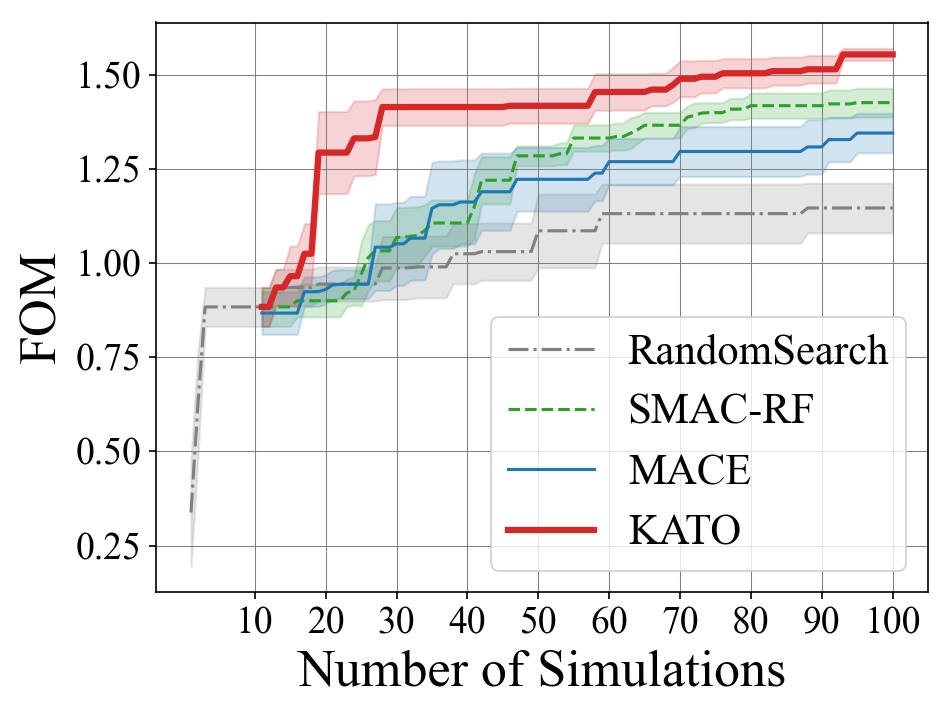}
\end{subfigure}
\begin{subfigure}[b]{0.29\linewidth}
    \caption{Bandgap}
    \label{Bandgap}
    \vspace{-0.04in}
    \includegraphics[width=\linewidth]{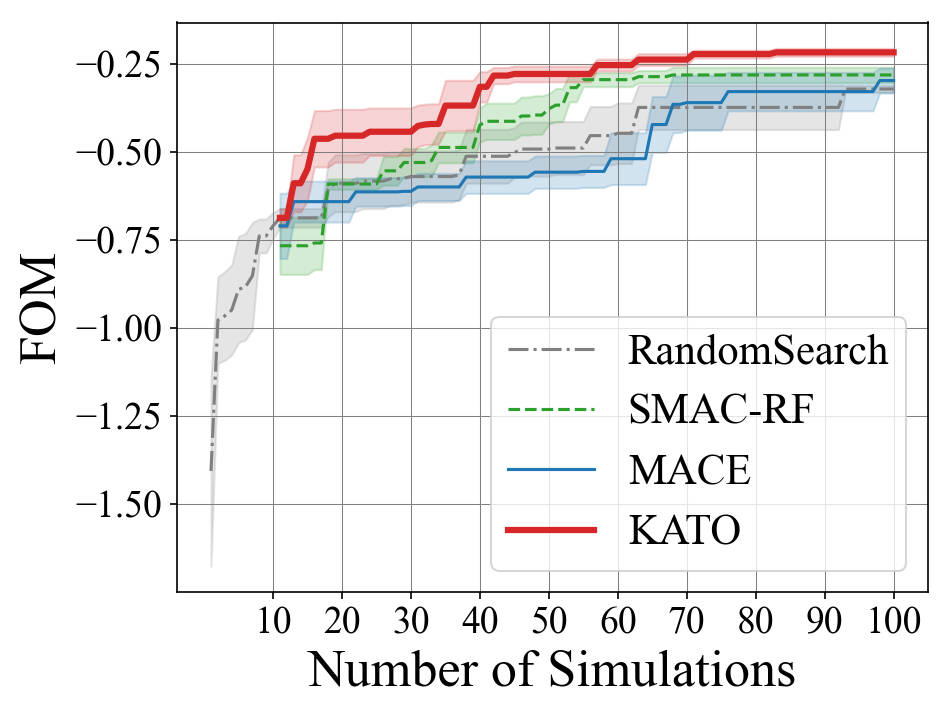}
\end{subfigure}
\vspace{-0.2in}
\caption{Transistor sizing by optimizing FOM}
\label{fig: fom}
\vspace{-0.1in}
\end{figure*}
\vspace{-0.15in}
\subsection{Assessment of FOM Optimization}
\vspace{-0.05in}
We initiated the evaluation based on the FOM of \Eqref{eq: FOM}.
\ours was compared against SOTA Bayesian Optimization (BO) techniques for a single objective, SMAC-RF\footnote{\href{https://github.com/automl/SMAC3}{https://github.com/automl/SMAC3}}, along with MACE and a naive random search (RS) strategy.
All methods are given 10 random simulations as the initial dataset, and the sizing results (FOM versus simulation budget) for the 180nm technology node are shown in \Figref{fig: fom}.
SMAC-RF is slightly better than MACE due to this simple single-objective optimization task.
Notably, \ours outperforms the baselines by a large margin.
Particularly, \ours consistently achieves the maximum FOM, with up to 1.2x improvement, and it takes about 50\% fewer simulations to reach a similar optimal FOM.
The optimal result of RS does not actually satisfy all constraints, highlighting the limitation of FOM-based optimization.

\cmt{
--
First, we assess the proposed method on the three circuits based on the FoM of \Eqref{eq: FoM}.
We compare \ours with the SOTA BO methods for single objective optimization, namely MACE\footnote{\href{https://github.com/automl/SMAC3}{https://github.com/automl/SMAC3}
}  and SMAC-RF~\cite{lindauer2022smac3}, and the naive random search (RS).
SMAC-RF is a golden standard for surrogate-based optimization by using a random forest and gradient boosting tree as the surrogate.
MACE is widely used in circuit design and transistor sizing due to its simplicity and effectiveness.
All methods are given the same random samples as the initial dataset. \YYwrite{We conducted 10 random samples for each model and then observed the values of FoM at convergence after 90 simulation cost.}The statistical FoM versus simulation budget over five runs is shown in \Figref{fig: fom}.
SMAC-RF is better than MACE due to the simplicity of the FoM function. However, \ours still outperforms SMAC-RF by a large margin, which demonstrates the effectiveness of the proposed method.
Particularly, with 100 simulations as budget, \ours always achieves the best outcome, with up to 1.3x improvement in final FoM and an average $50\%$ simulation cost to reach a similar optimal FoM.
}

\begin{figure*}
    \vspace{-0.2in}
    \centering
\begin{subfigure}[b]{0.29\linewidth}
    \caption{Two-stage OpAmp}
    \label{Two-stage operational amplifier}
    \vspace{-0.02in}
    \includegraphics[width=\linewidth]{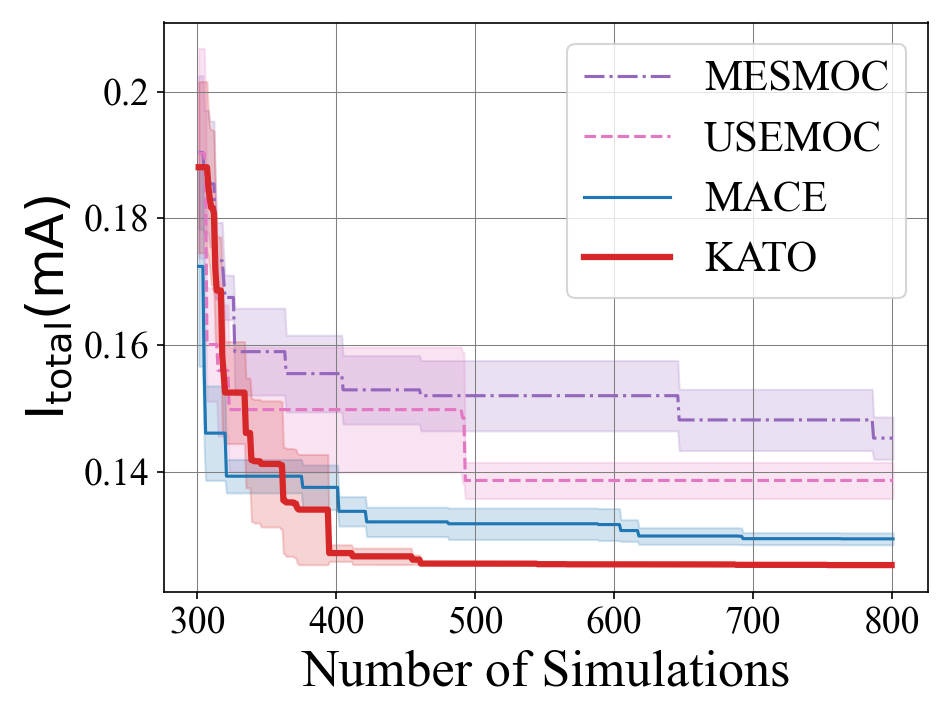}
\end{subfigure}
\begin{subfigure}[b]{0.29\linewidth}
    \caption{Three-stage OpAmp}
    \label{Three-stage operational amplifier}
    \vspace{-0.02in}    \includegraphics[width=\linewidth]{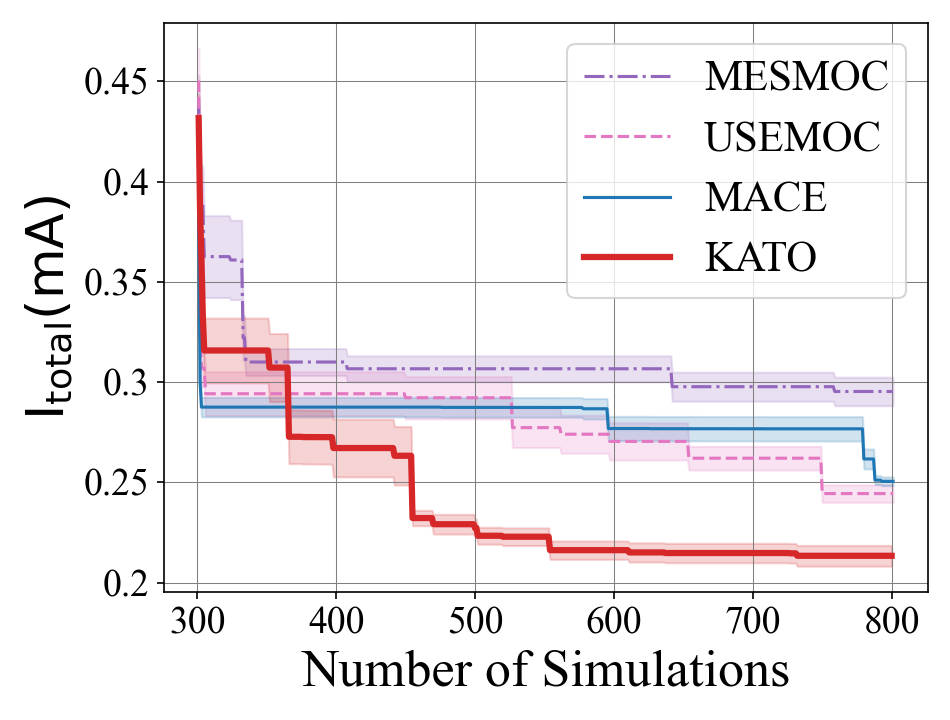}
\end{subfigure}
\begin{subfigure}[b]{0.29\linewidth}
    \caption{Bandgap}
    \label{Bandgap circuit}
    \vspace{-0.02in}
    \includegraphics[width=\linewidth]{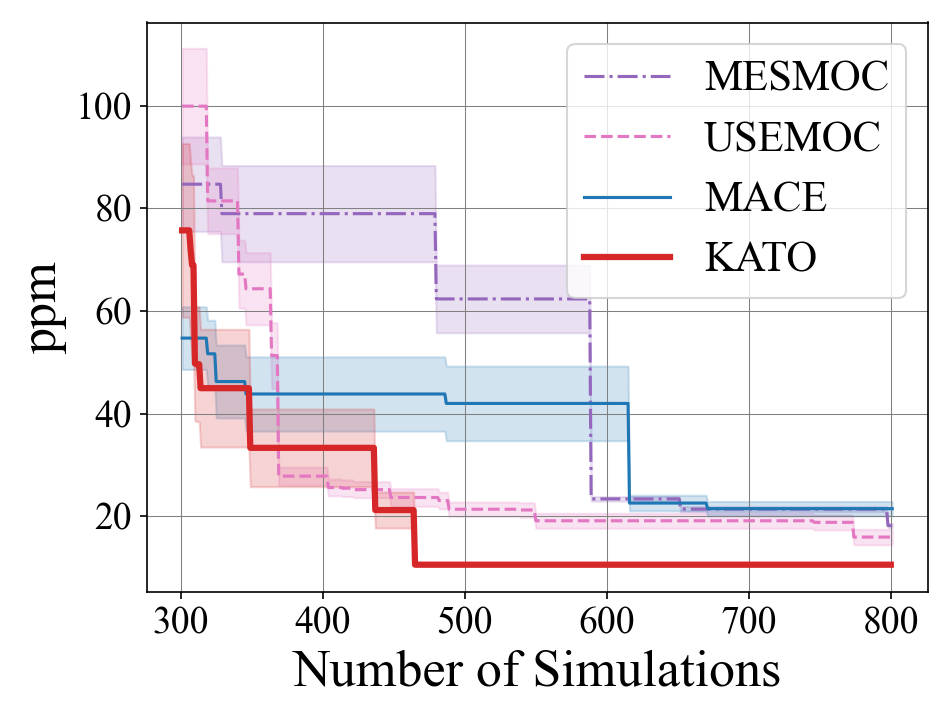}
\end{subfigure}
\vspace{-0.2in}
\caption{Transistor sizing by constrained optimization}
\label{fig: Constraint Optimization 180}
\vspace{-0.2in}
\end{figure*}
\vspace{-0.1in}

\begin{table*}
    \centering
        \tabcolsep=5pt
        \caption{Transistor Sizing Optimal Performance via Optimization with Constraints}
        \label{AllCaseTable}
        \vspace*{-0.15in}
        \begin{adjustbox}{width=2\columnwidth,center}
        \renewcommand\arraystretch{1.0}
        \begin{tabular}{l|cccc|cccc|ccc}
          \toprule[1.5pt]
          & \multicolumn{4}{|c|}{Two-stage OpAmp(180nm)} & \multicolumn{4}{|c|}{Three-stage OpAmp(180nm)}  & \multicolumn{3}{|c}{Bandgap(180nm)} \\
          \midrule
          Method & I(uA) & Gain(dB) & PM($^\circ$) & GBW(MHz) & I(uA)  & Gain(dB) & PM($^\circ$) & GBW(MHz)  & TC(ppm/$^\circ$C) & I(uA) & PSRR(dB @100Hz)\\
          \midrule[1pt]
          Specifications & min & >60 & >60 & >4 & min & >80 & >60 & >2 & min & <6 & >50\\
          Human Expert  &274.84 &75.12  &63.57&8.23 &462.84&112.49&65.61&2.05 & 11.26 & 5.31   & 61.01  \\
          MSEMOC & 138.58 & 80.39  & 65.28 & 5.66 & 288.28 & 86.85  & 73.55 &2.27 & 14.04 & 5.29   & 60.97 \\
          USEMOC & 137.89 & 65.42  &66.63& 4.63 & 230.47 & 81.24 & 73.67 & 2.09  & 10.36 &4.78 & 61.21\\
          MACE  & 127.69 & 79.30  &61.50&4.38 & 245.62 & 81.02 & 68.38 & 2.13 & 10.41 & 4.78 & 61.20 \\
          KATO & \bf{124.21} &61.18  & 60.59&4.56 & \bf{187.51} & 80.3  & 63.99 & 2.10 & \bf{9.66} & 5.42 & 61.99\\
        \bottomrule[1.5pt]
      \end{tabular}
      \vspace*{-0.5in}
      \end{adjustbox}
      \vspace*{-0.25in}
  \end{table*}

\vspace*{-0.05in}
\subsection{Assessment of Constrained Optimization}
\vspace{-0.05in}
Next, we assess the proposed method of transistor sizing with a more practical and challenging constrained optimization setup.
During optimization, only designs satisfying all constraints are considered valid and included in the performance reports. 
To provide sufficient valid designs for the surrogate model, we first simulate 300 random designs, typically yielding about 7 valid designs, a 2.3\% that makes RS not applicable in this task.

We compare \ours with SOTA constrained BO tailored for circuit design, namely, MESMOC~\cite{belakaria2020max}, USEMOC~\cite{belakaria2020uncertainty}, and MACE with constraints~\cite{zhang2021efficient}. 
%
The results of 180nm are shown in \Figref{fig: Constraint Optimization 180}, where MESMOC shows a poor performance due to its lack of exploration, and MACE is generally good except for the three-stage OpAmp.
\ours demonstrates a consistent superiority, always achieving the best performance with a clear margin, and most importantly, with about
50\% of simulation cost to reach the best-performing baseline.
The final design performance is shown in Table \ref{AllCaseTable}, where \ours achieves the best performance by extreme trade-off for the constraints (\eg Gain) as long as they fulfilling the requirements.

\begin{figure*}[ht]
\vspace{-0.0in}
\centering
\begin{subfigure}[c]{0.45\linewidth}
    \caption{Two-stage OpAmp (180nm) To Two-stage OpAmp (40nm)}
    \centering
    \vspace{-0.03in}
    \includegraphics[width=0.49\linewidth]{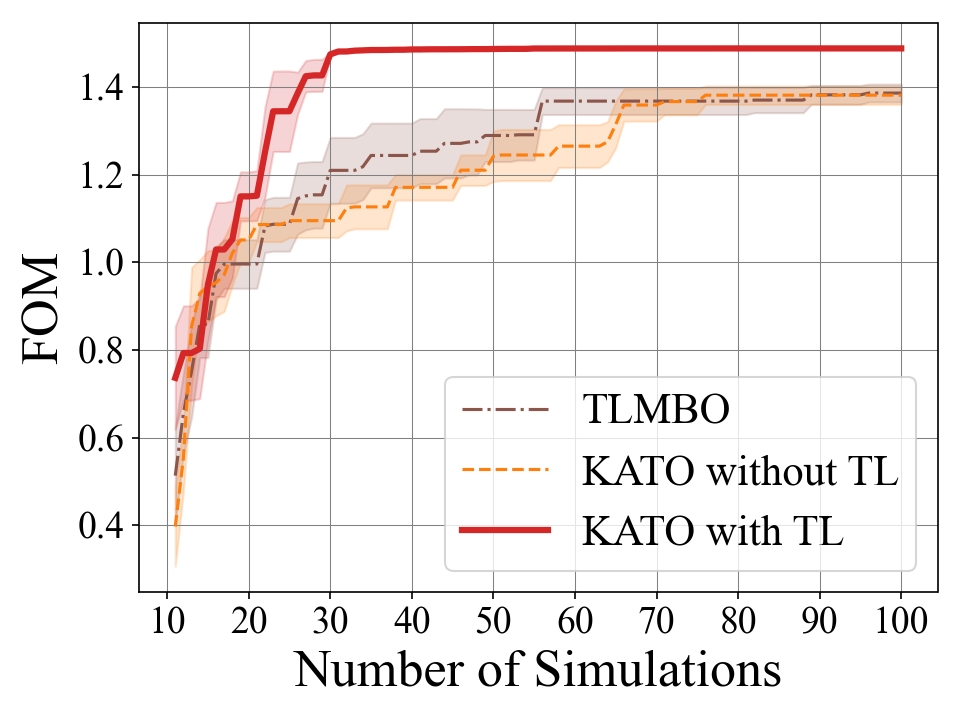}
    \includegraphics[width=0.49\linewidth]{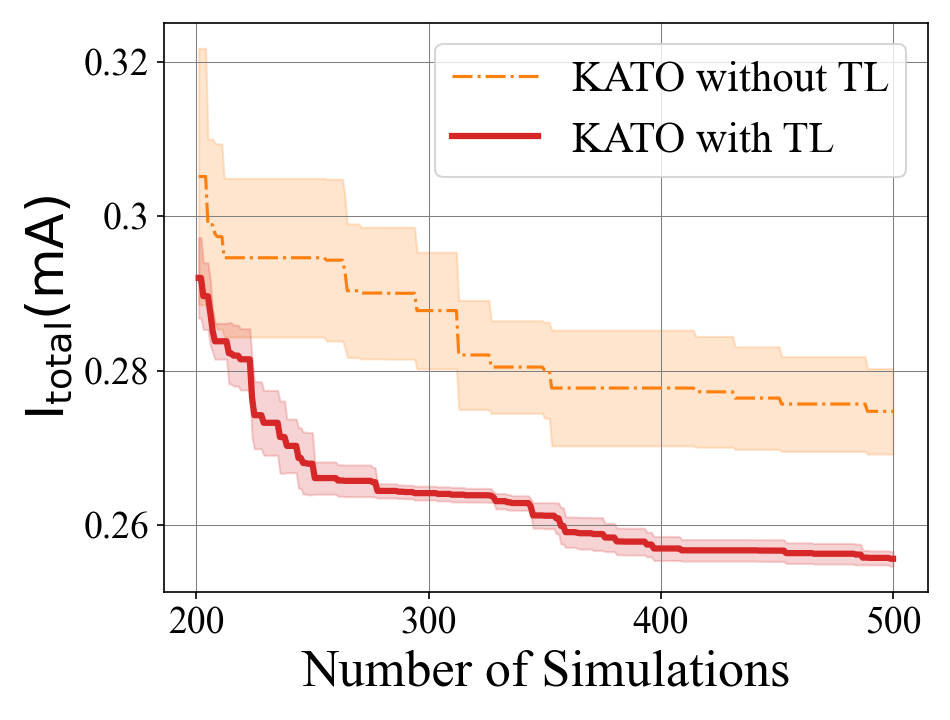}
\end{subfigure}
\begin{subfigure}[c]{0.45\linewidth}
    \caption{Three-stage OpAmp (180nm) To Three-stage OpAmp (40nm)}
    \vspace{-0.03in}
    \centering
    \includegraphics[width=0.49\linewidth]{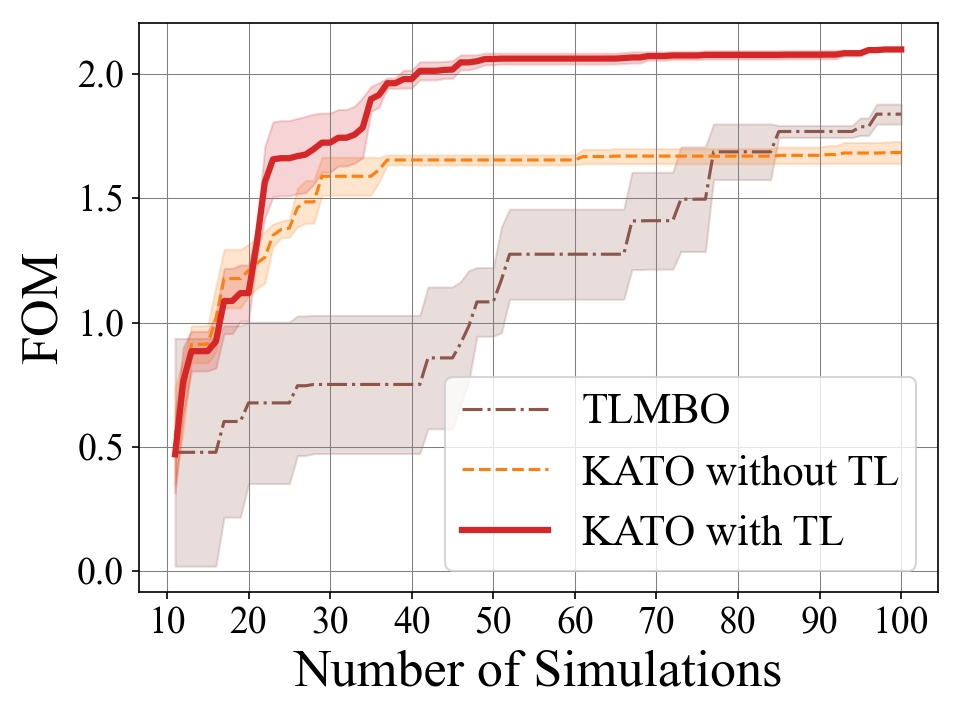}
    \includegraphics[width=0.49\linewidth]{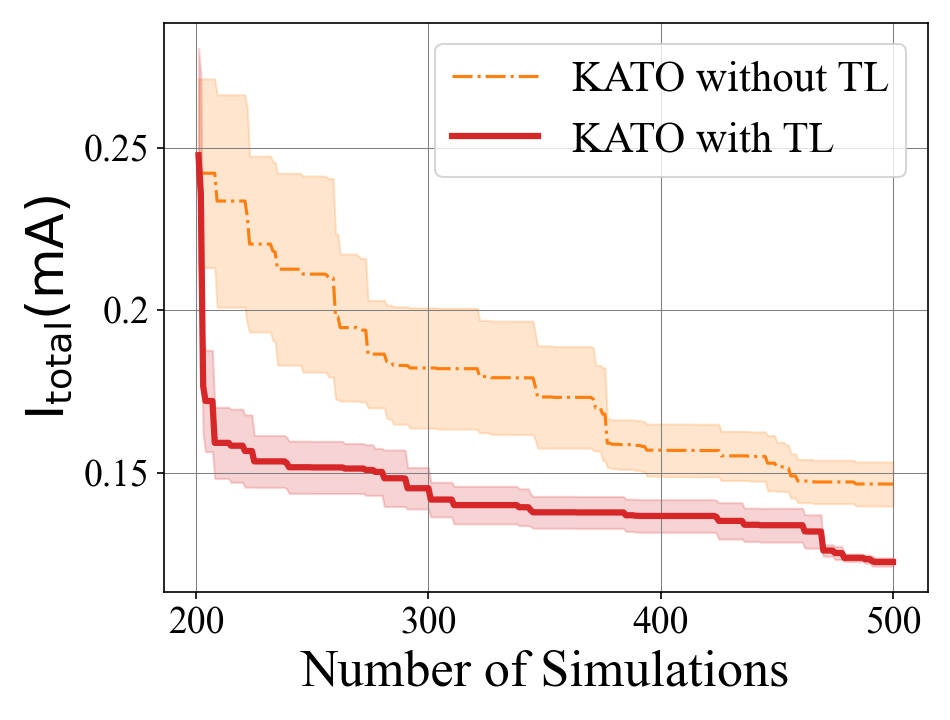}
\end{subfigure}\\
\vspace{-0.1in}
\begin{subfigure}[c]{0.45\linewidth}
    \caption{Three-stage OpAmp (40nm) To Two-stage OpAmp (40nm)}
    \vspace{-0.03in}
    \centering
    \includegraphics[width=0.49\linewidth]{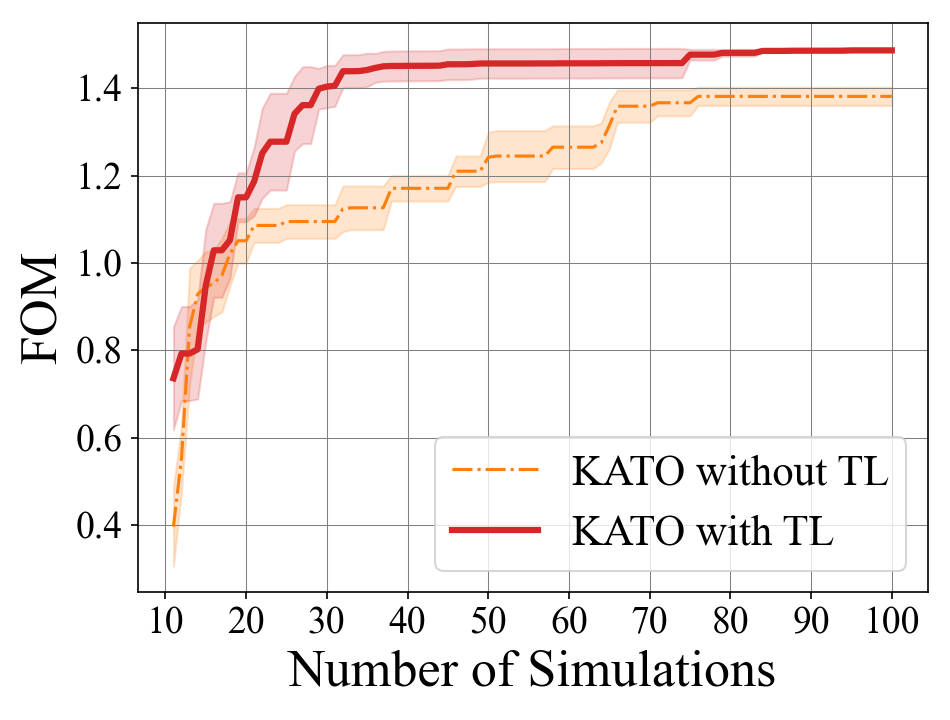}
    \includegraphics[width=0.49\linewidth]{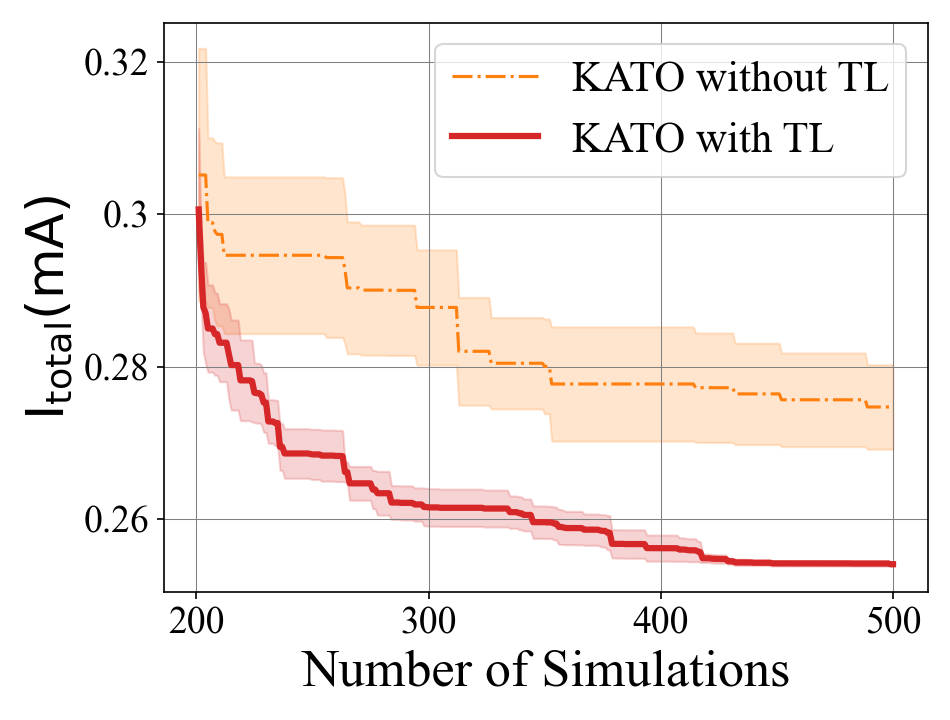}
\end{subfigure}
\begin{subfigure}[c]{0.45\linewidth}
    \caption{Two-stage OpAmp (40nm) To Three-stage OpAmp (40nm)}
    \vspace{-0.03in}
    \centering
    \includegraphics[width=0.49\linewidth]{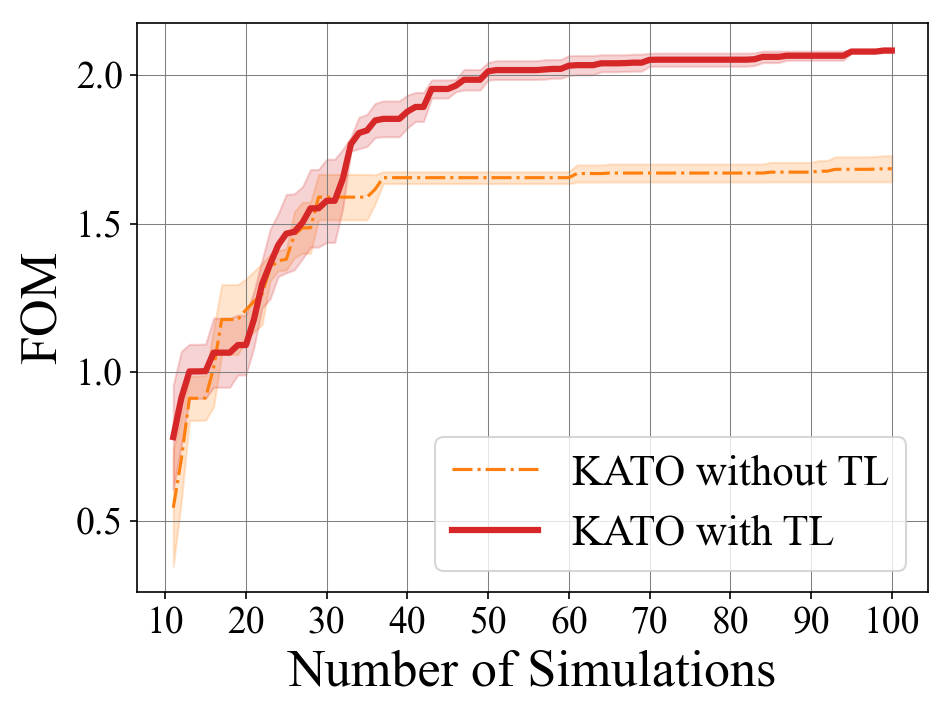}
    \includegraphics[width=0.49\linewidth]{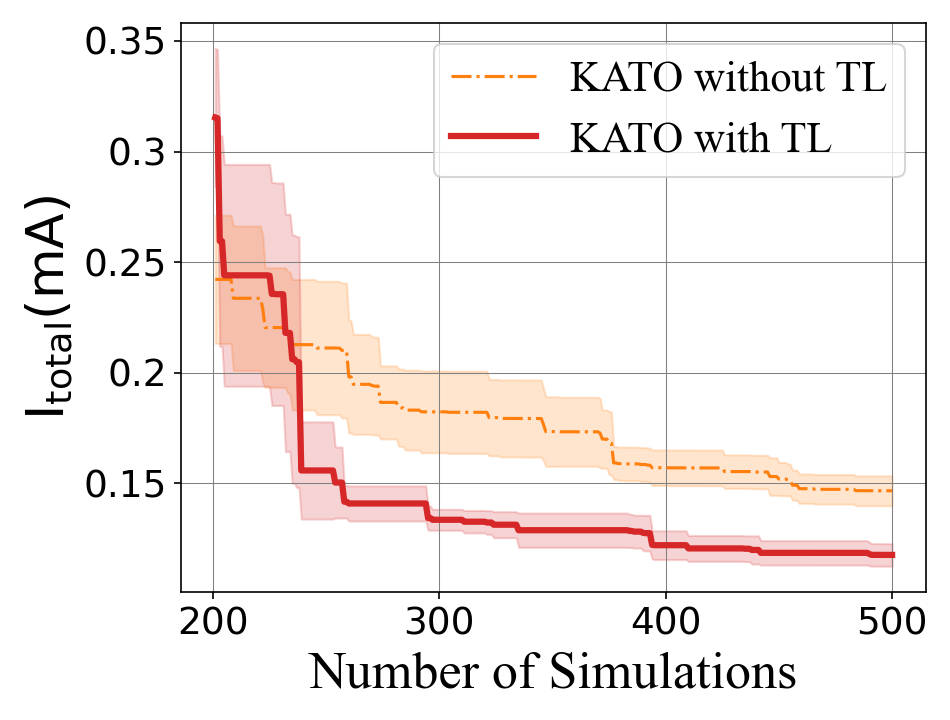}
\end{subfigure}
\\
\vspace{-0.1in}
\begin{subfigure}[c]{0.45\linewidth}
    \caption{Three-stage OpAmp (180nm) To Two-stage OpAmp (40nm)}
    \vspace{-0.03in}
    \centering
    \includegraphics[width=0.49\linewidth]{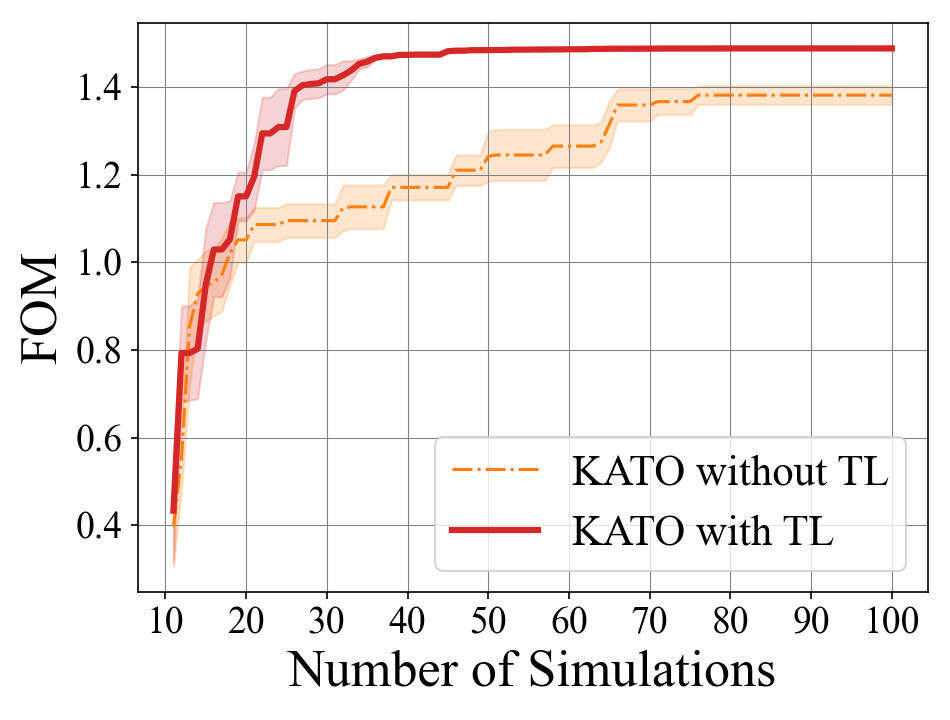}
    \includegraphics[width=0.49\linewidth]{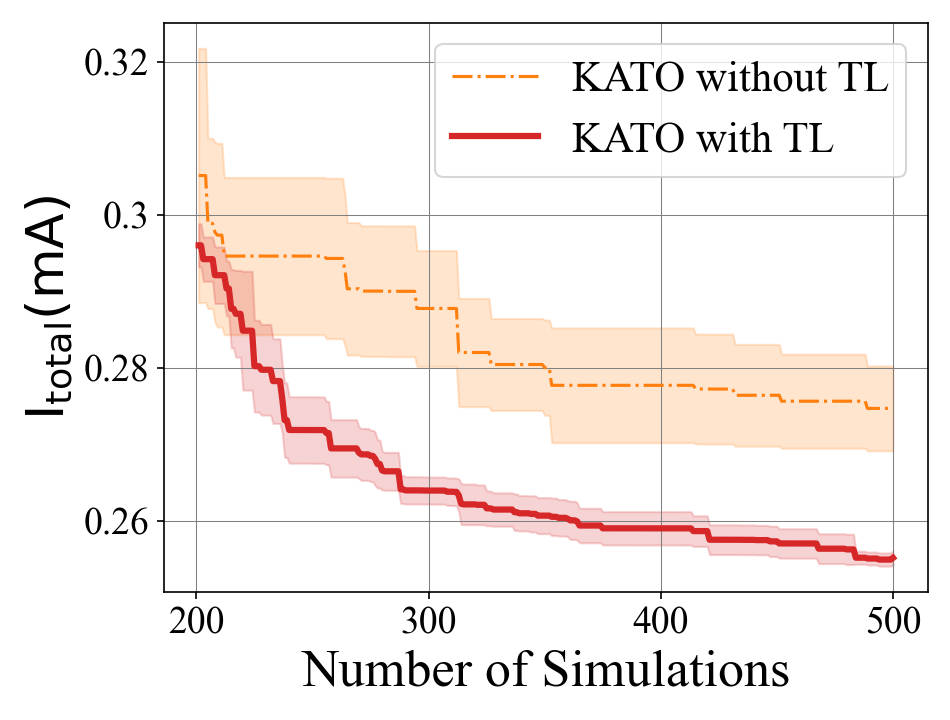}
\end{subfigure}
\begin{subfigure}[c]{0.45\linewidth}
    \caption{Two-stage OpAmp (180nm) To Three-stage OpAmp (40nm)}
    \vspace{-0.03in}
    \centering
    \includegraphics[width=0.49\linewidth]{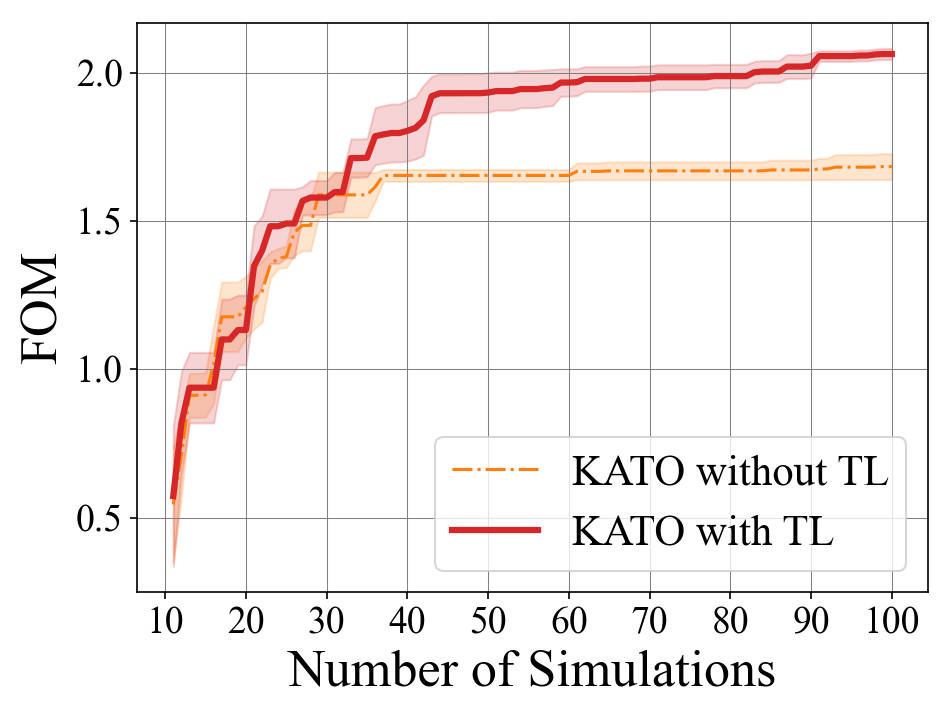}
    \includegraphics[width=0.49\linewidth]{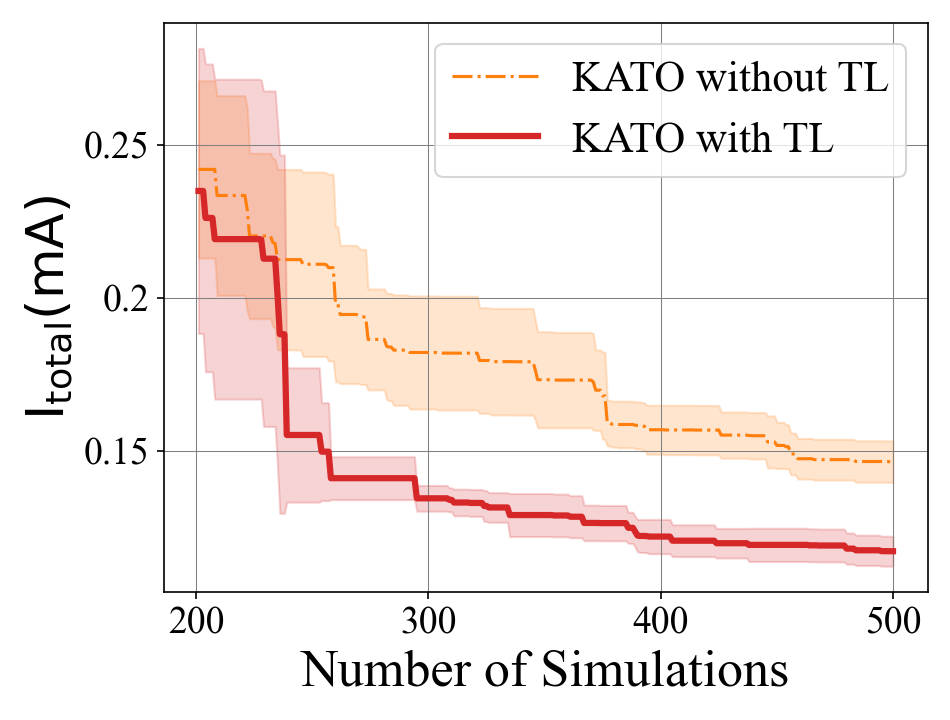}
\end{subfigure}
\vspace{-0.15in}
\caption{Transistor sizing constrained optimization with \tl of designs and technology node}
\label{fig: tl of node and design}
\vspace{-0.25in}
\end{figure*}

\vspace{-0.05in}
\subsection{Assessment of Transfer Learning}
\vspace{-0.05in}
Finally, we assess \ours with transfer learning between different topologies and technology nodes for FOM and constrained optimization.
Each experiment provides 200 random samples for the source data; the initial random target sample size is {10} for the FOM optimization and {200} for the constrained optimization.
We mainly conduct transfer leaning on the two-stage and three-stage OpAmps due to their similarity in topology and technology node. Note that the design variable is different.

\noindent\textbf{Transfer learning between technology nodes.}
We first compare \ours to the SOTA BO equipped with transfer learning, namely, TLMBO~\cite{zhang2022fast}
(which is only capable of FOM optimization). Other RL-based methods, \eg RL-GCN~\cite{wang2020gcn}, perform poorly due to the small data size setup of this experiment and thus are not included in the comparison. We also compared \ours with and without the transfer learning for the two-stage and three-stage OpAmps in the 40nm technology node.
The statistical results over five random runs are shown in 
Figs. \ref{fig: tl of node and design}(a) and \ref{fig: tl of node and design}(b), which highlights the effectiveness of the transfer learning, delivering an average 2.52x speedup (defined by the simulations required to reach the best performance of the \ours without \tl) and 1.18x performance improvement.

\noindent\textbf{Transfer learning between topologies.}
As far as the authors know, there is no BO method that can perform transfer learning between different topologies. We thus validate \ours with and without transfer learning between two-stage and three-stage OpAmps in the 40nm technology node.

The statistical results are shown in Figs. \ref{fig: tl of node and design}(c) and \ref{fig: tl of node and design}(d), which demonstrate the effectiveness of the transfer learning, delivering an average 2.35x speedup and 1.16x performance improvement.

\noindent\textbf{Transfer learning between topologies and technology nodes.}
Finally, we assess transfer learning with both topologies and technology nodes. The results are shown in Figs. \ref{fig: tl of node and design}(e) and \ref{fig: tl of node and design}(f), which demonstrates the effectiveness of the transfer learning, delivering an average 2.40x speedup and 1.16x performance improvement.

The final design performance is shown in Table \ref{tab: final2}. \tl between technology nodes achieves the best results as it is the easier task. Nonetheless, the difference between different transfer learning tasks is not significant. Compared to human experts in three-stage OpAmp, \ours shows up to 1.62x improvement in key performance.

\vspace*{-0.22in}
\begin{table}[H]
\centering
\caption{Transistor Sizing Optimal Performance via Optimization with Constraints with \tl}
\label{tab: final2}
\vspace*{-0.18in}
\begin{adjustbox}{width=0.95\columnwidth,center}
\renewcommand\arraystretch{1.0}
\setlength{\tabcolsep}{3pt}
\begin{tabular}{ll|ccccc}
\toprule[1.5pt]
\multicolumn{6}{c}{\textbf{Two Stage OpAmp(40nm)}} \\
\midrule
& Method & I(uA) & Gain(dB) & PM($^\circ$) & GBW(MHz) \\ 
\midrule
& Specifications & min & >50 & >60 & >4 \\
& Human Expert  &308.10 &51.77 &71.33 &7.08 \\
& KATO & 273.04 & 52.44  & 81.24 & 21.09 \\
& KATO (TL Node)  & \bf{254.05} & 50.29  & 83.72 & 15.05  \\
& KATO (TL Design) & {257.12}  & 50.04  & 82.68 & 10.28 \\
& KATO (TL Node\&Design) & {258.01} & 51.23  & 85.78 & 13.31\\
\midrule[1pt]
\multicolumn{6}{c}{\textbf{Three Stage OpAmp(40nm)}} \\
\midrule
& Specifications & min & >70 & >60 & >2 \\
& Human Expert  &244.72 &74.10&60.18&2.03 \\
& KATO & 151.09 & 70.23 & 69.85 & 3.49\\
& KATO (TL Node) & \bf{118.47} & 74.41 & 71.84 & 2.65 \\
& KATO (TL Design) & {118.71}  & 71.46 & 72.92 & 2.43 \\
& KATO (TL Node\&Design) & {120.08} & 70.44  & 73.44 & 2.48\\
\bottomrule[1.5pt]
\end{tabular}
\end{adjustbox}
\vspace{-.3in}
\end{table}

\vspace{-0.05in}
\section{Conclusion}
\vspace{-0.05in}
We propose, \ours, a novel transfer learning for transistor sizing, which enables transferring knowledge from different designs and technologies for BO for the first time. Except for improving the SOTA, we hope the idea of KAT can inspires more interesting research.
Further extension includes extending transfer learning to many different circuits of various types, e.g., SRAM, ADC, and PLL.

\vspace{-0.15in}
\bibliographystyle{ACM-Reference-Format}
\bibliography{KATO}

\appendix

\end{document}